\documentclass[lettersize,journal]{IEEEtran}
\usepackage{amsmath,amsfonts}
\usepackage{array}
\usepackage[caption=false,font=normalsize,labelfont=sf,textfont=sf]{subfig}
\usepackage{textcomp}
\usepackage{subcaption}
\usepackage{url}
\usepackage{verbatim}
\usepackage{graphicx}
\usepackage{tabularx} % 用于自动调整列宽的表格
\usepackage{cite}
\usepackage{float}    % 用于 [H] 强制位置选项
\usepackage{booktabs}
\usepackage{multirow}
\usepackage{graphicx}
\usepackage{xcolor}
\usepackage{bbding}
\usepackage{pifont}
\usepackage{makecell}
\usepackage{siunitx}
\usepackage{rotating}
\usepackage{amsmath}   % 提供 align、cases 等环境
\usepackage{amssymb}   % 提供 \mathbb 等黑板粗体
\usepackage{dblfloatfix}  % 在导言区

\usepackage[ruled,linesnumbered,boxed,noend,vlined]{algorithm2e}
\usepackage{amssymb}
\begin{document}

\title{Acetrans: An Autonomous Corridor-Based and Efficient\\ UAV Suspended Transport System}


\author{Weiyan Lu$\textsuperscript{‡}$, Huizhe Li$\textsuperscript{‡}$, Yuhao Fang, Zhexuan Zhou, Junda Wu, Yude Li, Youmin Gong$^{*}$, Jie Mei$^{*}$
        % <-this % stops a space
\thanks{‡ W. Lu and H. Li contribute equally to this work.}
\thanks{* Corrsponding authors: Jie Mei {\tt\small jmei@hit.edu.cn} and Youmin Gong {\tt\small gongyoumin@hit.edu.cn}.}
\thanks{The authors are with the School of Intelligence and Engineering, Harbin Institute of Technology, Shenzhen, Guangdong, China.}%}%
}


\IEEEpubid{}

\maketitle

\begin{abstract}
Unmanned aerial vehicles (UAVs) with suspended payloads offer significant advantages for aerial transportation in complex and cluttered environments. However, existing systems face critical limitations, including unreliable perception of the cable–payload dynamics, inefficient planning in large-scale environments, and the inability to guarantee whole-body safety under cable bending and external disturbances. This paper presents \textbf{Acetrans}, an \textbf{A}utonomous, \textbf{C}orridor-based, and \textbf{E}fficient UAV suspended \textbf{trans}port system that addresses these challenges through a unified perception, planning, and control framework. A LiDAR–IMU fusion module is proposed to jointly estimate both payload pose and cable shape under taut and bent modes, enabling robust whole-body state estimation and real-time filtering of cable point clouds. To enhance planning scalability, we introduce the \textbf{M}ulti-size-\textbf{A}ware \textbf{C}onfiguration-space \textbf{I}terative \textbf{R}egional \textbf{I}nflation (MACIRI) algorithm, which generates safe flight corridors while accounting for varying UAV and payload geometries. A spatio-temporal, corridor-constrained trajectory optimization scheme is then developed to ensure dynamically feasible and collision-free trajectories. Finally, a nonlinear model predictive controller (NMPC) augmented with cable-bending constraints provides robust whole-body safety during execution. Simulation and experimental results validate the effectiveness of Acetrans, demonstrating substantial improvements in perception accuracy, planning efficiency, and control safety compared to state-of-the-art methods.
\end{abstract}


\begin{table}[!t]
\caption{Nomenclature}
\centering
\begin{tabularx}{\columnwidth}{l >{\raggedright\arraybackslash}X}
$\mathcal{Q}, \mathcal{L}, \mathcal{W}, \mathcal{I}$& Quadrotor frame, payload frame,  world frame and IMU frame. \\
$\boldsymbol{e}_x, \boldsymbol{e}_y, \boldsymbol{e}_z \in \mathbb{R}^3$ & Unit vector in the axis-direction of the world frame. \\
$\boldsymbol{x}_B, \boldsymbol{y}_B, \boldsymbol{z}_B \in \mathbb{R}^3$ & Unit vector in the axis-direction of the quadrotor frame. \\[6pt]
$\boldsymbol{x}_\mathcal{I}, \boldsymbol{y}_\mathcal{I}, \boldsymbol{z}_\mathcal{I} \in \mathbb{R}^3$ & Unit vector in the axis-direction of the IMU frame. \\[6pt]
$\boldsymbol{x}_Q, \boldsymbol{x}_L \in \mathbb{R}^3$ & Position of the quadrotor and the payload. \\
$\boldsymbol{v}_Q, \boldsymbol{v}_L \in \mathbb{R}^3$ & Linear velocity of quadrotor and the payload. \\
${}^{\mathcal{W}}{R_{\mathcal{Q}}}, {}^{\mathcal{W}}{R_{\mathcal{I}}} \in \mathrm{SO(3)}$ & Orientation of the quadrotor, IMU in World frame. \\
$\boldsymbol{\omega}_Q \in \mathbb{R}^3$ & Body rate of the quadrotor. \\
$m_Q, m_L \in \mathbb{R}$ & Mass of the quadrotor and the payload. \\
$r_Q, r_L \in \mathbb{R}$ & Safe margin radius of the quadrotor and the payload. \\[6pt]

$\boldsymbol{J}_Q \in \mathbb{R}^{3\times 3}$ & Inertia matrix of the quadrotor. \\
$\boldsymbol{M}_Q \in \mathbb{R}^3$ & Control torque acting on the quadrotor. \\
$\boldsymbol{F}_Q \in \mathbb{R}^3$ & External force acting on the quadrotor. \\
$\boldsymbol{F}_L \in \mathbb{R}^3$ & External force acting on the payload. \\[6pt]

$\tilde{\boldsymbol{\rho}} \in \mathbb{R}^3$ & The tangent direction at the end of the rope points from the part near the quadrotor towards the payload. \\
$\boldsymbol{\rho} \in S^2$ & Unit vector in the direction of $\tilde{\boldsymbol{\rho}}$. \\
$\dot{\boldsymbol{\rho}} \in \mathbb{R}^3$ & First derivative w.r.t. time of vector $\boldsymbol{\rho}$. \\
$\boldsymbol{\omega}_\rho \in \mathbb{R}^3$ & Angular velocity of the cable direction $\boldsymbol{\rho}$. \\[6pt]

$l \in \mathbb{R}$ & Distance between the quadrotor and the payload. \\
$l_0 \in \mathbb{R}$ & Cable length. \\
$l_L \in \mathbb{R}$ & Payload length. \\
$f, f_T \in \mathbb{R}$ & Thrust of the quadrotor, and tension in the cable. \\
\end{tabularx}
\end{table}

\section{Introduction}
\IEEEPARstart{U}{nmanned} aerial vehicles (UAVs) have shown great promise in aerial transportation for logistics, agriculture, and disaster relief. Existing aerial transport platforms can be broadly categorized into fixed-wing UAVs, multirotors with active manipulators, and multirotors with suspended cables.

Fixed-wing UAVs \cite{10802785, 10587983} offer long range, high speed, and energy efficiency, making them suitable for large-scale coverage and long-distance delivery. However, they require runways or large open spaces for takeoff and landing, lack hovering capability, and exhibit limited maneuverability in cluttered environments. In addition, their payload handling is inflexible—loading and unloading cargo is difficult and often requires ground infrastructure. These limitations render fixed-wing UAVs unsuitable for agile, low-altitude transport in forests, urban canyons, or indoor scenarios, where precise and flexible operations are required.

Multirotors with active manipulators \cite{Peng2025, doi:10.1126/sciadv.adv6629, doi:10.1126/sciadv.adn6642} provide dexterity for rigidly grasping payloads, but this comes with increased mechanical complexity, higher system weight, reduced payload capacity, limited motion range due to kinematic constraints, higher deployment cost, and greater susceptibility to damage.

Multirotors with cable-suspended loads \cite{palunko2012Trajectory,sreenath2013TrajectoryGenerationControlc, foehn2017FastTrajectoryOptimization} strike a favorable balance. They require no additional actuation, reducing system weight and control dimensionality while enabling heavier payloads and simpler cargo handling. The cable suspension provides natural flexibility for attachment and release, lower cost, and reduced mechanical fragility compared with manipulators. In contrast to fixed-wing platforms, multirotors can hover, take off and land vertically without the need for runways, and operate safely in low-altitude, cluttered environments such as forests or urban canyons. Moreover, cable-suspended systems allow efficient load pick-up and drop-off by means of electromagnetic grippers \cite{foehn2017FastTrajectoryOptimization}, a method that is impractical for fixed-wing aircraft requiring ground infrastructure for payload handling. These properties make cable-suspended multirotors the most scalable and practical approach for real-world aerial transportation in complex and constrained environments.

Traditionally, suspended-load flights have been manually operated, which demands exceptional piloting skill. Ensuring safety at high speed is nearly impossible for human pilots, so manual suspended-load flights are usually restricted to open-air, high-altitude scenarios. This precludes operation in dense forests, urban areas, or other cluttered spaces where aerial transport would be most impactful. A fully autonomous suspended-load system thus represents a major advance: it enables UAVs to perform fast, safe, low-altitude payload transportation in complex environments that were previously inaccessible, with the capability of autonomous perception, planning, and control.

In the perception of suspended-load UAVs, whole-body perception is crucial for safe operation.
Unlike bare UAVs, a suspended system includes a payload and a cable, both of which must be localized to guarantee collision-free flight. The cable is particularly challenging: its thin, dynamic structure generates spurious point clouds that corrupt localization and mapping, and wind or rotor downwash can bend it, enlarging the system footprint. Most prior work assumes a taut cable, which may result in unsafe or infeasible trajectories.

Existing perception strategies remain limited.Vision-based approaches \cite{Tang2018Aggressive,ren2021vision,Fotis2023motion,Jocob2024Vision,Panetsos2024Aerial} offer rich environmental cues but suffer from narrow fields of view, blind spots around the payload and cable, and poor robustness in darkness. Many also presume the cable is always taut, preventing accurate estimation under disturbances. Disturbance-observer (DOB) schemes \cite{Lee2017Autonomous,Li2023Observer} bypass imaging but still rely on tension, losing accuracy when the cable bends, and they provide no awareness of surrounding obstacles—limiting their use in unknown environments.

In the planning and control of suspended-load UAVs, whole-body safety has been studied extensively. However, existing methods still face several limitations in achieving it. Some constrain only UAV or payload, leaving the cable unprotected~\cite{real_time_payload2019, impactor}. Some earlier methods \cite{Mixed_Kumar2015,foehn2017FastTrajectoryOptimization,payload_ral2020, Son2020Real, Potdar2020Online} avoid collisions by applying bounding geometric shapes or simpler obstacle shapes, though these tend to be conservative and not very effective when the scene becomes complex. To handle complex environments, some recently works (e.g., Autotrans~\cite{autotrans} and Impactor~\cite{impactor}) build the obstacle-avoidance strategy based on Euclidean Signed Distance Fields (ESDF). While effective in small-scale indoor settings, ESDF computation becomes extremely time-consuming in large-scale outdoor environments. In comparison, the construction of safe flight corridors (SFC) is computationally more efficient and triggered only before replanning, making corridor-based strategies potentially more advantageous. However, existing corridor-based planning methods~\cite{rils,gcopter,super} are primarily designed for bare UAVs. Due to the complex geometry of suspended-load systems, these methods cannot be directly applied and require tailored extensions to effectively handle such configurations.

Critically, while Impactor~\cite{impactor} first exploits cable-bending modes to enhance agility rather than explicitly guaranteeing safety,  even though bent cable deformation plays a critical role in ensuring collision safety. 

In addition, Most planning methods \cite{autotrans, foehn2017FastTrajectoryOptimization, KumarICRA2013load} neglect the effect of wind disturbances on cable direction, resulting in trajectories that are unsafe. 

Concluding the separate discussion on perception, planning and control, we observe that most existing works either focus solely on perception without incorporating autonomous obstacle-avoidance planning or control~\cite{Tang2018Aggressive,ren2021vision,Fotis2023motion,Jocob2024Vision,Panetsos2024Aerial, Lee2017Autonomous,Li2023Observer}, or rely on motion-capture-based planning and control while lacking autonomous state estimation~\cite{real_time_payload2019, autotrans, impactor, Mixed_Kumar2015,foehn2017FastTrajectoryOptimization,payload_ral2020, Son2020Real, Potdar2020Online}. Therefore, to date, there remains a lack of a fully autonomous suspended-load system that unifies perception, planning, and control.

% In summary, two fundamental gaps remain. First, there is a lack of robust whole-body perception that estimates both payload pose and cable shape under bending and disturbances. Second, there is an absence of planning and control frameworks that rigorously guarantee whole-body safety in cluttered environments.

In summary, research on UAV suspended-load systems still faces three major limitations:  
1) there is a lack of robust whole-body perception capable of estimating both the payload pose and the cable shape under bending and external disturbances, with reliable performance under varying illumination conditions and in both indoor and outdoor environments;
2) there is an absence of planning and control frameworks that rigorously guarantee whole-body safety, for both taut and bent cable, in unstructured environments; 
3) there is no unified perception–planning–control framework enabling fully autonomous transportation of suspended loads at any time, day or night, in both indoor and outdoor environments.

To overcome the above limitations, we introduce an \textbf{A}utonomous, \textbf{C}orridor-based and \textbf{E}fficient UAV suspended \textbf{trans}port system (\textbf{Acetrans}), which to the best of our knowledge represents the first fully autonomous suspended-load framework that unifies perception, planning, and control, and achieves safe and efficient payload transportation around the clock in complex indoor and outdoor environments. 

On the hardware level, we employ a dual-LiDAR configuration (Section~\ref{sec:system setup}) to independently perceive obstacles around both the multirotor and the payload. In parallel, we propose a LiDAR-IMU fusion framework (Section~\ref{sec:whole-body perception}) for high-frequency payload pose estimation, which also enables accurate high-rate fitting of the cable under both taut and bending conditions. Based on this fitting, we remove cable and payload points from the point cloud to prevent interference with localization and mapping. Combined with Fast-LIO2~\cite{Xu2022fastlio2} for UAV state estimation, this forms the first unified perception and estimation framework that simultaneously supports the UAV, cable, and payload. Benefiting from LiDAR-based sensing, the system operates reliably under all-day conditions, regardless of lighting. 

To fill the gap in corridor-based planning for suspended-load systems, we propose a tailored corridor generation strategy (Section~\ref{sec:maciri}) named Multi-size-Aware Configuration-space Iterative Regional Inflation (MACIRI) and a robust initialization scheme (Section~\ref{sec:trans_init}) that avoids infeasible trajectories. At the planning level, we adopt the Minimum Control Effort (MINCO) optimization framework~\cite{gcopter}  and leverage the inclusion property of convex hulls (Section~\ref{sec:property_corr}) to significantly enhance both the safety and computational efficiency of whole-body obstacle avoidance. This allows our system to handle complex, unstructured 3D environments, narrow spaces, and thin obstacles while ensuring whole-body safety, achieving optimization speeds 1--3 orders of magnitude faster than existing baselines (Section~\ref{sec:experiment}). 

At the control level, to ensure fundamental safety for suspended-load systems under cable-bending conditions, we design a novel NMPC framework that integrates the generated safe corridors and introduces conservative trapezoidal constraints capable of enclosing the bent cable (Section~\ref{sec:nmpc_bend}). This approach fills the gap in obstacle avoidance for cable-bending scenarios. Overall, our \textbf{Acetrans} system represents a milestone in achieving fully autonomous suspended-load aerial transportation, unifying perception, planning, and control into a single, robust framework.

The main contributions are summarized as follows: 

\begin{enumerate}
    % \item \textbf{Whole-body perception for suspended-load systems:} To the best of our knowledge, we present the first perception framework that jointly estimates both payload pose and cable shape under disturbances. A Lidar-IMU based module enables accurate payload pose estimation across indoor and outdoor environments, independent of illumination conditions. In addition, we design a disturbance-aware cable-shape estimator based on a rotated catenary model, which reconstructs cable geometry at high frequency. 
    
    \item \textbf{Whole-body perception for suspended-load systems:} We propose a LiDAR--IMU framework that estimates the payload pose and surrounding environment in both indoor and outdoor settings, independent of lighting, and fits the cable curve to remove its points from the map, improving localization and mapping. To the best of our knowledge, this is the first whole-body perception framework for suspended-load systems.

    \item \textbf{Multi-size-Aware Configuration-space Iterative Regional Inflation (MACIRI):} We propose a corridor generation strategy that employs seeds with varying vertex sizes, ensuring complete seed coverage while maximally expanding the corridor volume to better capture feasible configurations for suspended-load systems.
    
    \item \textbf{Corridor-based planning for suspended-load systems:}  
    We develop a spatio-temporal and corridor-based trajectory optimization framework for suspended-load systems. The method guarantees whole-body safety while remaining computationally efficient, achieving an improvement in optimization speed by 1--3 orders of magnitude.

    \item \textbf{Cable-bending-safe control for suspended-load systems:}  
    We propose a control framework implemented through an corridor-based NMPC scheme augmented with trapezoidal cable-bending envelopes. By simultaneously constraining the UAV, payload, and bending cable, the controller guarantees robust whole-body safety in the presence of disturbances, and, to the best of our knowledge, this is the first framework that achieves obstacle avoidance under cable bending.

\end{enumerate}

\section{Related Work}

\subsection{Pose Estimation of Suspended Loads}
A variety of methods have been proposed for payload perception and state estimation.

Disturbance-observer (DOB) techniques \cite{Lee2017Autonomous} estimate external forces from IMU and load-cell data to infer swing angles, but they fundamentally depend on cable tension and fail once the cable bends.

Vision-based approaches dominate the literature. Tang et al.\cite{Tang2018Aggressive} fused fisheye-camera and IMU data in an EKF using the tether constraint, while Ren et al.\cite{ren2021vision} combined monocular imagery with an encoder-based sensor. 
Other works used supervised learning \cite{Fotis2023motion}, geometric fitting \cite{Jocob2024Vision}, or visual servoing for anti-swing control \cite{Yu2025Visual}. 
Although effective under taut conditions and good lighting, these methods degrade in darkness or when the cable bends.
Panetsos et al.\cite{Panetsos2024Aerial} employed an event camera with Bézier fitting, enabling load observation in low light, yet still assuming a taut cable and offering limited performance in complete darkness.

However, as no existing work addresses perception under bent-cable conditions, there is likewise no suspended-load UAV system that estimates and exploits the cable shape when it is deformed.

To handle darkness and bent-cable scenarios, we propose a LiDAR-IMU fusion framework. LiDAR actively emits laser beams and provides precise depth measurements for 3-D payload localization, while IMU data improve estimation and supply the tangent direction at the cable end. 
Together, these measurements allow fitting the full cable shape, which is further used to filter out points belonging to the cable and payload from the LiDAR map, preventing them from corrupting localization and mapping.

\subsection{Generating Free Convex Polytope}
\label{section:corridor_related_work}
Constructing free convex polytopes is a fundamental technique for safe and efficient robotic navigation. By enclosing a seed region within convex polytopes that separate free space from obstacles, collision-avoidance constraints can be expressed as linear inequalities. This abstraction simplifies trajectory optimization, improves computational tractability, and facilitates online planning in cluttered environments. Larger convex polytopes further provide more usable space for trajectory search, enabling smoother and more dynamically feasible paths. The key challenges lie in balancing polytope quality, computational efficiency, and guaranteed seed inclusion.

Existing approaches to account for robot size in polytope construction can be broadly classified into three categories.  
The first is point inflation~\cite{ren_online,Faster,fuzhangSwarm}, where obstacle points are inflated by the robot’s radius before convex decomposition. While conceptually simple, this method suffers from exponential growth in the number of points and often generates small, sharp polytopes that over-constrain optimization.  
The second is plane shrinkage~\cite{rils,liu2023integrated,elastic_tracker}, which constructs polytopes directly from obstacle points and then shrinks them by the robot’s radius. Although it avoids point inflation, this method often produces overly conservative regions, and shrinking after convergence may even exclude the seed itself, disrupting corridor continuity.  
The third is sphere modeling, where obstacle points are directly represented as spheres with radii equal to the robot’s size. This representation eliminates the need for inflation, improves computational efficiency, and generally produces larger polytopes compared with shrinkage-based methods.

Building upon these foundations, iterative frameworks have been developed to improve both polytope quality and efficiency. Early on, Deits et al.~\cite{iris} proposed the Iterative Regional Inflation by Semidefinite Programming (IRIS) algorithm, which pioneered iterative inflation by monotonically increasing the volume of the maximum-volume inscribed ellipsoid (MVIE) of the convex polytope at each iteration. However, IRIS \cite{iris} accepts only point seeds and does not guarantee seed containment. To overcome these shortcomings, Liu et al.~\cite{rils} proposed the Regional Inflation by Line Search (RILS) algorithm, which introduces line segments as seeds. It first generates a maximal ellipsoid that contains the seed while excluding obstacles, and then inflates it in the same manner as IRIS \cite{iris}. Nevertheless, for suspended-load systems, the seeds may take the form of lines, quadrilaterals, or tetrahedra, which are derived from the system's geometry and front-end path planning. These seed types cannot be accommodated by either IRIS\cite{iris} or RILS\cite{rils}.  

More recently, Wang et al.~\cite{firi} proposed Fast Iterative Regional Inflation (FIRI), which introduced Restrictive Inflation (RsI) to ensure seed containment and present an equivalent second-order conic programming (SOCP) formulation for MVIE to improve computational efficiency, extending seed types to convex sets. FIRI~\cite{firi} can be divided into two variants: FIRI with point inflation (FIRI-Inflation) and FIRI with plane shrinkage (FIRI-Shrinkage). Ren et al.~\cite{super} subsequently introduced the Configuration-space Iterative Regional Inflation (CIRI) algorithm, which extends FIRI~\cite{firi} by utilizing sphere modeling to generate free convex polytopes. Compared with FIRI-Inflation and FIRI-Shrinkage, CIRI improves corridor volume and seed containment, as demonstrated in drone racing scenarios~\cite{super}. However, CIRI’s corridor containment correction is primarily designed for line-segment seeds and cannot robustly handle general convex-hull seeds, making it unsuitable for suspended-load systems. More importantly, all of the above algorithms, including IRIS, RILS, FIRI, and CIRI, consider only a single robot size and thus fail to fully exploit the available free space for multi-size robotic systems, such as suspended-load transportation, where the system is described by geometric seeds (e.g., lines or convex-hulls) and the size of each vertex varies. For example, these methods must be executed using the maximum size of either the UAV or the payload, leading to overly conservative corridor construction.

To address these limitations, our MACIRI framework enables safe corridor generation for suspended-load UAV systems with varying UAV–payload size ratios. Unlike prior single-size formulations that must conservatively adopt the maximum size of either the UAV or the payload and thus fail to fully exploit available free space, MACIRI models obstacles as adaptive spheres based on the nearest seed vertex size, effectively enlarging usable corridor volumes. Moreover, MACIRI generalizes corridor generation from point and line seeds to arbitrary convex-hull seeds and introduces a robust containment correction strategy that guarantees $100\%$ enclosure of all seed vertices. These capabilities allow MACIRI to produce larger, less conservative, and feasible safe flight corridors, making it particularly well-suited for suspended-load aerial transport systems operating in cluttered environments.

\subsection{Whole-body Planning and Control of UAV with Suspended Payload}
For UAVs with suspended payloads, many approaches have been proposed for whole-body planning, considering either whole-body safety or the kinematic and dynamic constraints of the entire system. Palunko et al. \cite{palunko2012Trajectory} proposed a dynamic programming-based trajectory generation method to suppress payload oscillations during agile maneuvers, but it relied on predefined waypoints and lacked explicit obstacle avoidance. Sreenath et al. \cite{KumarICRA2013load} modeled the system as a differentially flat hybrid system and minimized load snap to reduce oscillations but also lacked obstacle avoidance.  

Subsequent research began to explicitly incorporate obstacle avoidance into trajectory planning. Son et al. \cite{real_time_payload2019} employed a convexification-based obstacle avoidance strategy using A* and a “project-and-linearize” approach to convert non-convex collision constraints into convex ones, with MPC refining the resulting trajectory to ensure dynamic feasibility. However, their obstacle constraints are imposed only on the payload position, and obstacles are restricted to cylinders. 

Furthermore,\cite{Mixed_Kumar2015,foehn2017FastTrajectoryOptimization,payload_ral2020, Son2020Real, Potdar2020Online} computed the relative relationship between the whole body and the obstacles simplified as basic geometric shapes (e.g., spheres, cylinders or cuboids) to perform whole-body obstacle avoidance.
 Despite these advancements, they share common limitations: obstacles are simplified and their positions are assumed to be known a priori, which limits their applicability in complex environments.
 
To handle complex scenarios, Autotrans~\cite{autotrans} and Impactor~\cite{impactor} were proposed, which primarily model obstacles using grid maps instead of simplified geometric shapes, and optimize trajectories for obstacle avoidance based on an ESDF map. However, the computation of the ESDF map is expensive, making it challenging to achieve real-time performance on low-computation platforms or in large-scale environments. Regarding whole-body obstacle avoidance, Impactor~\cite{impactor} ignores the cable and only constrains the multirotor and the payload, whereas Autotrans~\cite{autotrans} achieves whole-body avoidance by applying obstacle constraints on sampled points along the cable. The drawback of Autotrans~\cite{autotrans} is that its computation time increases with the cable length or sampling resolution, and the sampling-based strategy cannot theoretically guarantee the safety of unsampled points on the cable. For Impactor~\cite{impactor}, although it improves agility by optimizing complementary constraints between taut and bent cable modes, it does not consider the safety of the cable in bent conditions, and the high computational cost makes real-time execution challenging.

In contrast to these methods, we adopt a corridor-based planning and control strategy tailored to suspended-load systems, particularly suitable for complex or large-scale environments. Without sampling cables and constructing ESDF maps, our approach naturally provides enhanced whole-body safety while maintaining higher computational efficiency. Furthermore, we introduce a  NMPC controller that explicitly accounts for cable-bending obstacle avoidance, thereby providing robust whole-body safety and filling a significant gap in existing planning and control frameworks for suspended-load UAV systems.

\section{Preliminaries}
\subsection{Nominal System Dynamics}
\label{sec:nominalsystemdynamics}
The dynamic behavior of a quadrotor system carrying a cable-suspended payload can be categorized into two primary operational modes based on the tension state of the cable: (i) a taut mode, characterized by continuous tension in the cable resulting in coupled motion between the quadrotor and payload, and (ii) a bent mode, wherein the cable becomes untensioned, leading to decoupled dynamics and free-fall motion of the payload.
In line with the approach adopted in \cite{KumarICRA2013load,KumarCDC2013load}, the system dynamics can be characterized under two distinct modes, as detailed below.
\subsubsection{Taut-cable dynamics}
\begin{subequations}
\begin{align}
\dot{\boldsymbol{x}}_{L} &= \boldsymbol{v}_{L}, \label{eq:taut_xL}\\
(m_Q+m_L)\bigl(\dot{\boldsymbol{v}}_{L}+g\boldsymbol{e}_z\bigr)
&= \Bigl(\boldsymbol{\rho}\!\cdot\! f{}^{\mathcal{W}}{R_{\mathcal{Q}}}\boldsymbol{e}_z
- m_Q l_0\,\dot{\boldsymbol{\rho}} \cdot \dot{\boldsymbol{\rho}}\Bigr)\boldsymbol{\rho}
\; \nonumber\\ &\qquad +\;\boldsymbol{F}_{Q}+\boldsymbol{F}_{L}, \label{eq:taut_trans}\\
\dot{\boldsymbol{\rho}} &= \boldsymbol{\omega}_{\rho}\times \boldsymbol{\rho}, \label{eq:taut_rho_kin}\\
m_Q l\,\dot{\boldsymbol{\omega}}_{\rho} &= -\,\boldsymbol{\rho}\times \bigl(f{}^{\mathcal{W}}{R_{\mathcal{Q}}}\boldsymbol{e}_z + \boldsymbol{F}_Q - \boldsymbol{F}_L\bigr), \label{eq:taut_rho_dyn}\\
\boldsymbol{x}_{Q} &= \boldsymbol{x}_{L}-l_0\,\boldsymbol{\rho}, \label{eq:taut_geom}\\
\dot{{}^{\mathcal{W}}{R_{\mathcal{Q}}}} &= {}^{\mathcal{W}}{R_{\mathcal{Q}}}\hat{\boldsymbol{\omega_Q}}, \\
\boldsymbol{J}_{Q}\dot{\boldsymbol{\omega_Q}} + \boldsymbol{\omega_Q}\times \boldsymbol{J}_Q\boldsymbol{\omega_Q}
&= \boldsymbol{M_Q}. \label{eq:taut_att}
\end{align}
\end{subequations}

\subsubsection{bent-cable dynamics}
\begin{subequations}
\begin{align}
\dot{\boldsymbol{x}}_{L} &= \boldsymbol{v}_{L}, \label{eq:slack_xL}\\
m_L\bigl(\dot{\boldsymbol{v}}_{L}+g\boldsymbol{e}_z\bigr) &= \boldsymbol{F}_{L}, \label{eq:slack_load}\\
\dot{\boldsymbol{x}}_{Q} &= \boldsymbol{v}_{Q}, \label{eq:slack_xQ}\\
m_Q\bigl(\dot{\boldsymbol{v}}_{Q}+g\boldsymbol{e}_z\bigr) &= f{}^{\mathcal{W}}{R_{\mathcal{Q}}}\boldsymbol{e}_z + \boldsymbol{F}_{Q}, \label{eq:slack_quad}\\
\dot{{}^{\mathcal{W}}{R_{\mathcal{Q}}}} &= {}^{\mathcal{W}}{R_{\mathcal{Q}}}\hat{\boldsymbol{\omega_Q}}, \\
\boldsymbol{J_Q}\dot{\boldsymbol{\omega_Q}} + \boldsymbol{\omega_Q}\times \boldsymbol{J_Q}\boldsymbol{\omega_Q}
&= \boldsymbol{M_Q}. \label{eq:slack_att}
\end{align}
\end{subequations}

For external forces, we directly use the method of Li et al. \cite{autotrans} to solve $\boldsymbol{F_Q}$ and $\boldsymbol{F_L}$.

\subsection{Differential Flatness}
\label{sec:flatness}

Differential flatness is a powerful structural property that allows a system's full state and control inputs to be expressed as algebraic functions of a set of flat outputs and a finite number of their derivatives. This representation significantly reduces the dimension of the trajectory optimization problem, enabling more efficient motion planning and control.

This work focuses exclusively on trajectory generation under the taut-cable mode. Although bent-cable dynamics can theoretically be modeled using hybrid systems with complementarity constraints, such formulations introduce significant numerical challenges, including poor convergence properties and increased computational overhead \cite{impactor}. In addition, bent-cable trajectory optimization complicates the incorporation of whole-body obstacle avoidance constraints. This is because obstacle-aware planning would require explicit estimation of the cable’s flexible shape at every time step, dramatically increasing the problem’s dimensionality and computational cost. Previous methods that considered bent-cable scenarios typically neglected such constraints, making them unsuitable for safety-critical applications \cite{impactor}.
In addition, under bent-cable modes, the payload is no longer directly controllable by the quadrotor and may exhibit free-fall behavior. This severely undermines the controllability of the system and introduces a disconnect between planned trajectories and their feasibility in execution. Even if a trajectory is successfully optimized under bent-cable assumptions, the lack of control authority during bent phases may lead to significant tracking errors or even collisions in cluttered environments.

In contrast, when the cable remains taut, the payload is fully controllable through the quadrotor, and the system adheres to differential flatness. This enables the generation of dynamically feasible and fully trackable trajectories with obstacle avoidance guarantees. While prior bent-cable methods were motivated by aggressive maneuvers such as passing through narrow gaps, similar agility can be achieved under taut-cable conditions by leveraging high-acceleration flight, which additionally ensures better control authority and safety during execution. Therefore, we choose to plan a trajectory in the fully controllable taut-cable mode and not actively trigger the incompletely controllable bent-cable mode.

After comparing the taut-cable mode and the bent-cable mode, differential flatness holds under the assumption that the cable remains taut throughout the motion. In this regime, the system behaves as a coupled multi-body structure, where the cable continuously transmits forces between the quadrotor and the payload. The flat outputs are given by $\mathcal{Z} = [\boldsymbol{x}_{L}, \psi]$, where $\boldsymbol{x}_{L}$ denotes the payload position and $\psi$ is the yaw angle of the quadrotor~\cite{KumarICRA2013load}. For simplicity and computational efficiency, the external forces $\boldsymbol{F}_Q$ and $\boldsymbol{F}_L$ are assumed to be constant during planning, such that their derivatives are $\boldsymbol{0}$.

The coupled dynamics can then be expressed as:

\begin{subequations}
\begin{align}
    \label{equ:flat_rho_ext_a}
    m_{L} \left( \ddot{\boldsymbol{x}}_{L} + g\boldsymbol{e}_{z} \right) &= -f_{T} \boldsymbol{\rho} + \boldsymbol{F}_L, \\
    \label{equ:flat_rot_ext_b}
    \boldsymbol{\rho} &= -\frac{ \ddot{\boldsymbol{x}}_{L} + g\boldsymbol{e}_{z} - \frac{1}{m_L} \boldsymbol{F}_L }{ \left\| \ddot{\boldsymbol{x}}_{L} + g\boldsymbol{e}_{z} - \frac{1}{m_L} \boldsymbol{F}_L \right\|_2 } \\
        \label{equ:flat_rot_ext_c}
    f{}^{\mathcal{W}}{R_{\mathcal{Q}}}\boldsymbol{e}_{z} &= m_{Q} \left( \ddot{\boldsymbol{x}}_{Q} + g\boldsymbol{e}_{z} \right) - f_{T} \boldsymbol{\rho} + \boldsymbol{F}_Q.
\end{align}
\end{subequations}

Equation~\eqref{equ:flat_rho_ext_a} models the payload dynamics, explicitly incorporating the effect of external forces $\boldsymbol{F}_L$ in addition to gravitational and cable forces. 
Equation~\eqref{equ:flat_rot_ext_b} defines the unit vector $\boldsymbol{\rho}$ along the cable direction, corrected by the external force $\boldsymbol{F}_L$ acting on the payload. 
Compared to standard quadrotor systems, the cable direction in a suspended-load system is more susceptible to external disturbances such as wind or contact forces. This sensitivity arises because $\boldsymbol{\rho}$ serves as a critical intermediate variable in reconstructing the quadrotor's thrust direction and trajectory from the flat outputs. Any deviation in $\boldsymbol{\rho}$ directly propagates to the computed control inputs. Ignoring $\boldsymbol{F}_L$ may thus lead to significant errors in the flatness-based trajectory, reducing accuracy and potentially destabilizing the control system.
Equation~\eqref{equ:flat_rot_ext_c} describes how the total thrust from the quadrotor must compensate not only for its own inertia and gravity but also for the reaction force from the cable and any disturbance force $\boldsymbol{F}_Q$.

Given the flat outputs $\boldsymbol{x}_{L}$ , $\psi$ , $\boldsymbol{F_Q}$ and $\boldsymbol{F_L}$, the quadrotor’s position $\boldsymbol{x}_{Q}$ and its derivatives can be obtained through Equation~\eqref{eq:taut_geom}. The thrust direction ${}^{\mathcal{W}}{R_{\mathcal{Q}}}\boldsymbol{e}_{z}$  and the thrust magnitude $f$ can be reconstructed using Equations~\eqref{equ:flat_rho_ext_a} , \eqref{equ:flat_rot_ext_b} and \eqref{equ:flat_rot_ext_c}, considering both internal coupling and external disturbances. With $\boldsymbol{x}_{Q}$, $\psi$ , and their derivatives available, the remaining states—such as angular velocity, full orientation and torques can be derived using conventional differential flatness-based methods~\cite{minimum_snap}.

By including external force terms, this formulation enhances robustness and enables more accurate trajectory generation in the presence of disturbances, thereby improving real-world applicability for aggressive or uncertain aerial manipulation tasks.

\subsection{Corridor Inclusion Constraints}
\label{sec:property_corr}
We study the inclusion property of corridors under linear inequality constraints.  
Let $A \in \mathbb{R}^{m \times n}$ and $b \in \mathbb{R}^m$, and define the admissible set
\begin{equation}
\mathcal{C} = \{\boldsymbol{x} \in \mathbb{R}^n \mid A\boldsymbol{x} \le b\}.
\end{equation}

\subsubsection{Line-Seed Corridor.}
Consider a corridor represented by a line segment between two endpoints 
$\boldsymbol{x}_a$ and $\boldsymbol{x}_b$.  
Every point on the segment is
\begin{equation}
\boldsymbol{x}(t) = t\boldsymbol{x}_a + (1-t)\boldsymbol{x}_b, 
\qquad t \in [0,1].
\end{equation}
If both endpoints satisfy
\begin{equation}
A\boldsymbol{x}_a \le b, \qquad A\boldsymbol{x}_b \le b,
\end{equation}
then every point on the segment also satisfies
\begin{equation}
A\boldsymbol{x}(t) = t A\boldsymbol{x}_a + (1-t) A\boldsymbol{x}_b \le b.
\end{equation}
Thus, for line-seed corridors, if both endpoints satisfy the constraints, then every point on the line segment is automatically feasible.  

\subsubsection{Convex-Hull-Seed Corridor.}
More generally, let the corridor be a convex hull 
$\mathcal{H} = \operatorname{Conv}(\{\boldsymbol{v}_1,\dots,\boldsymbol{v}_K\})$ 
with vertices $\boldsymbol{v}_k$.  
Any point in $\mathcal{H}$ can be expressed as
\begin{equation}
\boldsymbol{x} = \sum_{k=1}^K \lambda_k \boldsymbol{v}_k, 
\qquad \lambda_k \ge 0,\; \sum_{k=1}^K \lambda_k=1.
\end{equation}
If all vertices satisfy $A\boldsymbol{v}_k \le b$, then
\begin{equation}
A\boldsymbol{x} = \sum_{k=1}^K \lambda_k A\boldsymbol{v}_k 
\le \sum_{k=1}^K \lambda_k b = b,
\end{equation}
implying $\boldsymbol{x}$ is feasible.  
Similarly, for convex-hull-seed corridors, feasibility of all vertices implies that every interior point satisfies the constraints as well.  
Therefore, it suffices to impose the constraints only on the vertices.  

\subsection{System Overview}
% The overall framework of the proposed approach is depicted in Figure~\ref{fig:framework}.
The overall framework of the proposed approach is as follows.
The whole-body perception module estimates both the pose of the suspended load and the configuration of the cable, which can operate under either a taut-cable or a bent-cable mode. A kinodynamic A* algorithm is employed to compute a collision-free path, serving as a seed for the MACIRI algorithm. Following this, a SFC is constructed, within which the proposed waypoint initialization strategy assigns waypoints located in the overlapping regions of adjacent polytopes. This ensures that the initial trajectory is not only short but also entirely enclosed within the corridor, accounting for both the UAV and the suspended load. Based on these initialized waypoints, a spatio-temporal, corridor-constrained trajectory optimization method is applied to generate a smooth, dynamically feasible, and collision-free trajectory under the fully controllable taut-cable mode. 

The resulting polynomial trajectory is tracked using a NMPC. Furthermore, to address safety concerns associated with cable bending, the NMPC is augmented with constraints derived from a bent-cable shape estimator. Specifically, an estimator of the bent cable shape generates conservative trapezoidal regions that are guaranteed to fully enclose the estimated bent shape of the cable. These trapezoids are embedded within the polytopic structure of the flight corridor, ensuring that the UAV and load trajectories remain safely bounded even under significant cable deformation.

\section{Whole-Body Perception For UAV Suspended Transport Systems}
\label{sec:whole-body perception}
\subsection{Pose Estimation of Suspended Loads}
\subsubsection{Frame Design}

To address the limitations of vision-based sensors in low-light or nighttime conditions, and to enable simultaneous perception of both the payload and surrounding obstacles, we employ a LiDAR sensor instead of a camera for payload sensing. In addition, a high-precision 9-axis IMU is mounted at the end of the cable. This IMU configuration allows us not only to obtain accurate acceleration and angular velocity measurements for pre-integration, but also to estimate the cable-end orientation more reliably. 

To simplify the model, and because the payload attitude itself is neither the focus of our estimation nor of our control, we make the following assuption:

\textbf{\textit{Assumption 1}}: The payload attitude is identical to the cable-end attitude. 
  
This assumption is valid in the absence of significant impulsive forces and can be further guaranteed by reinforcing the attachment at the cable end. Even if the assumption is occasionally violated, our method can still recover a reasonably accurate estimate of the cable-end position, which is the quantity required for the subsequent estimation steps.

As shown in the figure, our current hardware configuration enables the estimation of the payload state $\left( \boldsymbol{x}_{L_k}, \boldsymbol{v}_{L_k} \right)$ and cable orientation $ {}^W{R_{I_k}}$ when the rope is taut, as well as the observation of the payload position and the cable-end orientation when the rope is bent. Furthermore, the overall state of the cable can be fitted from these observations, which can then be utilized in subsequent planning, including filtering out cable points in the mapping and enabling avoidance of bent cables in the control module.

\subsubsection{The Cable is Taut}
When the cable is taut, it is assumed to form a straight line. This critical assumption enables us to fuse multiple sensor sources within a unified optimization framework for a robust and high-fidelity state estimation of the payload. The framework is built upon two independent measurement models and the payload's kinematic constraints.

\paragraph{Measurement Models}
First, from the cable-end IMU, the constant direction vector along the cable corresponds to the IMU's orientation measurement ${}^\mathcal{W}{\hat{R}_{\mathcal{I}}}$. First, from the cable-end IMU, the constant direction vector along the cable 
corresponds to the IMU's orientation, ${}^\mathcal{W}{\hat{R}_{\mathcal{I}}}$. Without loss of generality, 
we assume the cable is aligned with the IMU's local z-axis. The direction vector 
$\boldsymbol{\rho}$ in the world frame is therefore given by:
\begin{equation}
\label{eq:rho_from_rotation}
\boldsymbol{\rho} = {}^\mathcal{W}{\hat{R}_{\mathcal{I}}} \begin{bmatrix} 0 \\ 0 \\ 1 \end{bmatrix}.
\end{equation}
The payload's position, $\boldsymbol{x}_L$, can then be determined from the known cable length $l_0$ and the quadrotor's position $\boldsymbol{x}_Q$, yielding the IMU direction measurement:
\begin{equation}
\label{imu_dir_factor}
\boldsymbol{x}^{\text{imu}}_L = \boldsymbol{x}_Q + \boldsymbol{\rho} \cdot (l_0 + \frac{l_L}{2} ) .
\end{equation}

Second, we obtain a position measurement from the downward-facing LiDAR. Inspired by Swarm-LIO2 \cite{fuzhangSwarm}, we attach reflective strips to the payload. The detection pipeline begins after receiving a new LiDAR scan. The raw points are undistorted \cite{fastlio1}, yielding a point cloud ${}^{\mathcal{Q}}\mathcal{P}$ in the quadrotor's body frame. To reduce computational load and mitigate interference, only points within a near-range threshold are processed. Points with high reflectivity are then efficiently grouped by Fast Euclidean Clustering (FEC) \cite{cao2022fec} to isolate the cluster originating from the payload and compute its centroid, ${}^{\mathcal{Q}}\breve{\boldsymbol{p}}_L$. This yields the LiDAR position measurement:
\begin{equation}
\label{lidar_factor}
\boldsymbol{x}^{\text{lidar}}_L = \boldsymbol{x}_Q + {}^\mathcal{W}{R_\mathcal{Q}} \cdot {}^{\mathcal{Q}}\breve{\boldsymbol{p}}_L .
\end{equation}

\paragraph{Factor Graph Optimization}
The two measurement models provide independent but complementary observations. We fuse these multimodal measurements (IMU kinematics, LiDAR position, and IMU orientation) with the system's geometric constraints within a factor graph framework, similar to the approach used LIO-SAM \cite{liosam2020shan}, to jointly estimate its full state trajectory. First, we define the full state vector $\mathcal{S}_k$ of the payload at each timestep $k$ within a sliding window of size $n+1$:
\begin{align}
\mathcal{S} &= \{ \boldsymbol{S}_0, \boldsymbol{S}_1, \dots, \boldsymbol{S}_n \}, \\
\boldsymbol{S}_k &= \left( \boldsymbol{x}_{L_k}, \boldsymbol{v}_{L_k}, {}^\mathcal{W}{R_{\mathcal{I}_k}}, \boldsymbol{b}_{a_k}, \boldsymbol{b}_{g_k} \right),
\end{align}
where the terms correspond to the payload's position, velocity, orientation of the cable end, and IMU biases.Note that based on \textbf{Assumption 1}, the orientation of the cable end, ${}^W{R_{I_k}}$, is also considered to be the orientation of the payload itself.

The Maximum a Posteriori (MAP) estimation of the trajectory $\mathcal{S}$ is then obtained by solving the following nonlinear weighted least-squares problem, which now includes four distinct factor types:
\begin{align} \label{eq:full_factor_optimization_weighted}
\min_{\mathcal{S}} \bigg\{ 
 \sum_{k=0}^{n-1} \| \boldsymbol{r}_{\mathcal{I}_k} \|^2_{\boldsymbol{\Sigma}_{\mathcal{I}}^{-1}} \nonumber 
& + \sum_{j \in \mathcal{C}} h \!\left( \| \boldsymbol{r}^{\text{lidar}}_j \|^2_{\boldsymbol{\Sigma}_{\text{lidar}}^{-1}} \right) \nonumber + \\
 \sum_{m \in \mathcal{D}} h \!\left( \| \boldsymbol{r}^{\text{imu\_pos}}_m \|^2_{\boldsymbol{\Sigma}_{\text{imu\_pos}}^{-1}} \right) \nonumber
& + \sum_{m \in \mathcal{D}} h \!\left( \| \boldsymbol{r}^{\text{imu\_ori}}_m \|^2_{\boldsymbol{\Sigma}_{\text{imu\_ori}}^{-1}} \right) 
\bigg\},
\end{align}

where the individual residual terms are defined as:
\begin{itemize}

    \item The IMU pre-integration factor residual, $\boldsymbol{r}_{\mathcal{I}_k}$, connects two consecutive states $\boldsymbol{S}_k$ and $\boldsymbol{S}_{k+1}$ based on the IMU's kinematic measurements, with noise covariance $\boldsymbol{\Sigma}_{\mathcal{I}}$.
    
    \item The LiDAR measurement factor residual, based on the measurement model in Eq. \eqref{lidar_factor}:
    \begin{equation}
        \boldsymbol{r}^{\text{lidar}}_j = \boldsymbol{x}_{L_j} - \boldsymbol{x}^{\text{lidar}}_{L_j}, 
    \end{equation}
    $\quad \boldsymbol{\Sigma}_{\text{lidar}}$ is the LiDAR noise covariance.

    \item The IMU Geometric Position Factor residual: This factor provides a geometric constraint linking the drone's position, the cable-end orientation, and the cable length to the payload's position.
        \begin{equation}
        \boldsymbol{r}^{\text{imu\_pos}}_m = \boldsymbol{x}_{L_m} - \boldsymbol{x}^{\text{imu}}_{L_m},
    \end{equation}
    where $\boldsymbol{\Sigma}_{\text{imu\_pos}}$ is its corresponding noise covariance.
    
    \item The IMU Orientation Factor residual: This factor directly constrains the estimated cable-end orientation against the orientation measurement provided by the IMU. The residual is defined on the SO(3) manifold as:
    \begin{equation}
        \boldsymbol{r}^{\text{imu\_ori}}_m = \text{Log}\left( \left({}^\mathcal{W}\hat{R}_{\mathcal{I}_m}\right)^T \cdot {}^\mathcal{W} R_{\mathcal{I}_m} \right),
    \end{equation}
    where ${}^\mathcal{W}\hat{R}_{I_m}$ is the direct orientation measurement from the IMU, ${}^\mathcal{W} R_{I_m}$ is the orientation state to be optimized, and $\text{Log}(\cdot)$ is the logarithmic map from SO(3) to its Lie algebra $\mathfrak{so}(3)$. $\boldsymbol{\Sigma}_{\text{imu\_ori}}$ is the IMU orientation noise covariance.
    
    \item $h(\cdot)$ represents the Huber robust cost function to reduce sensitivity to measurement outliers. $\mathcal{C}$ and $\mathcal{D}$ are the sets of available LiDAR and IMU direction measurements, respectively.
\end{itemize}

\subsubsection{The Cable is bent}

In the case of a bent cable, the cable may bend, causing the tangent direction at various points to change. Therefore, the IMU-measured direction at the cable's end can no longer be directly used to estimate the position of the payload.

Accordingly, in the factor graph, we remove the factor on the payload position that is derived from the straight-cable assumption in the IMU direction factor. However, since the IMU orientation is still highly accurate, it is retained as a factor representing the cable’s end orientation. Both the IMU pre-integration and LiDAR observations are independent of the straight-cable assumption, and thus these factors are also preserved. Furthermore, when new LiDAR measurements are not yet available, IMU observations are utilized to increase the update frequency. 

Finally, with the known cable length, we can fit the bent cable.

\subsubsection{Determination of Taut and Bent Cable Modes}

The cable mode is determined by evaluating the consistency between two independent payload position observations: 
the IMU-inferred position $\boldsymbol{x}^{\text{imu}}_L$ and the LiDAR-measured position $\boldsymbol{x}^{\text{lidar}}_L$, 
both of which have been introduced in Section~X. 
The discrepancy is defined as
\begin{equation}
    \delta \boldsymbol{x}_L = \boldsymbol{x}^{\text{imu}}_L - \boldsymbol{x}^{\text{lidar}}_L ,
\end{equation}
with its magnitude denoted as $\|\delta \boldsymbol{x}_L\|$.

Let $\tau_d$ be the discrepancy threshold and $\tau_\Sigma$ be the covariance threshold. 
The classification of change strategy is then performed as follows:

\begin{itemize}
    \item \textbf{Taut to bent:} The cable is regarded as bent if 
    $\|\delta \boldsymbol{x}_L\| > \tau_d$ or the position covariance from factor graph optimization satisfies 
    $\mathrm{tr}(\boldsymbol{\Sigma}_{x_L}) > \tau_\Sigma$ for at least $N$ consecutive frames.
    
    \item \textbf{bent to Taut:} The cable is regarded as taut if 
    $\|\delta \boldsymbol{x}_L\| \leq \tau_d$ for at least $N$ consecutive frames.
\end{itemize}

Here, $\boldsymbol{\Sigma}_{x_L}$ denotes the covariance of the estimated payload position obtained from factor graph optimization, 
and $N$ is the minimum number of consecutive frames required to avoid spurious switching due to measurement noise. 
This hysteresis-based formulation ensures a robust and stable transition between the taut and bent cable modes.

\subsection{Disturbance-Aware Estimation of the Bent Cable Shape}
\label{sec:cable_shape_etimator}
\subsubsection{Motivation.}
Estimating the 3D cable shape under external disturbances is essential in aerial load transportation systems.
\begin{itemize}
    \item \textbf{Point cloud filtering:} Cable point clouds are highly dynamic and interfere with localization and obstacle mapping. 
    To avoid being misinterpreted as obstacles and degrading SLAM performance, cable point clouds must be filtered at very high frequency.
    \item \textbf{Safety in bent-cable cases:} Current methods cannot guarantee safety when the cable bends, which makes motion planning difficult and unreliable. 
    For control purposes, cable shape estimation must run at extremely high frequency, 
    so that a disturbance-aware controller can react in real time and ensure obstacle avoidance when the cable bends.
\end{itemize}

\subsubsection{Catenary Model and the Effect of Wind on Cable Shape.}
Under the influence of gravity, a uniform cable with constant linear density $\lambda$ forms a catenary curve described by the equation:
\begin{equation}
z(x) = a \cosh\left( \frac{x}{a} \right),
\end{equation}
where $a = \frac{H}{\lambda g}$, with $H$ representing the horizontal tension and $g$ the gravitational acceleration. This equation describes the cable's shape in a uniform gravitational field.

\subsubsection{Wind and Gravity Fields.}
Assume that the wind field exerts a force on the cable similar to the gravitational force. For an infinitesimally small cable segment with constant mass per unit length, the forces due to wind and gravity are of equal magnitude and parallel. Let $\boldsymbol{g}_{\text{wind}}$ represent the acceleration due to the wind field and $\boldsymbol{g}_{\text{gravity}}$ represent the acceleration due to gravity. The total resultant acceleration field $\boldsymbol{g}$ is then the sum of the two fields:

\begin{equation}
\boldsymbol{g} = \boldsymbol{g}_{\text{gravity}} + \boldsymbol{g}_{\text{wind}}.
\end{equation}

This combined acceleration field determines the shape of the cable, which remains a catenary, but its orientation is defined by the direction of the resultant field $\boldsymbol{g}$.
 The resultant $\boldsymbol{g}$ remains uniform in magnitude and direction, which means the cable will still follow a catenary shape. However, instead of being aligned vertically with the gravitational field, the cable is now aligned with the direction of the resultant field $\boldsymbol{g}$.

The catenary curve under the combined effect of wind and gravity is thus the same in form as the classical catenary, but it is oriented according to the resultant acceleration field.

Consequently, we conclude that the cable shape under the combined effects of wind and gravity is still a catenary, but embedded in a plane aligned with the resultant acceleration field $\boldsymbol{g}$.

\subsubsection{Catenary in a Uniform Resultant Field.}
The solution to the catenary equation is:

\begin{equation}
z(x) = a \cosh\!\left(\frac{x - c}{a}\right) + C, \ z'(x) = \sinh\!\left(\frac{x - c}{a}\right).
\end{equation}
Setting the UAV attachment point at the origin $(0, 0)$, we have:
\begin{equation}
z(x) = a \left[ \cosh\!\left(\frac{x - c}{a}\right) - \cosh\!\left(\frac{c}{a}\right) \right],
\end{equation}
and the arc length to $x$ is:

\begin{equation}
L(x; a, c) = a \sinh\!\left(\frac{x - c}{a}\right) + a \sinh\!\left(\frac{c}{a}\right).
\end{equation}

\subsubsection{3D-to-2D Reduction by Rotation.}
Given UAV $\boldsymbol{p}_u$, load $\boldsymbol{p}_\ell$ and observed tangent $\boldsymbol{d}_\ell$, we construct an orthonormal basis
\begin{equation}
\hat{\boldsymbol{x}}=\tfrac{\boldsymbol{p}_\ell-\boldsymbol{p}_u}{\|\boldsymbol{p}_\ell-\boldsymbol{p}_u\|},\quad
\hat{\boldsymbol{z}}=\tfrac{(\boldsymbol{p}_\ell-\boldsymbol{p}_u)\times\boldsymbol{d}_\ell}{\|(\boldsymbol{p}_\ell-\boldsymbol{p}_u)\times\boldsymbol{d}_\ell\|},\quad
\hat{\boldsymbol{y}}=\hat{\boldsymbol{z}}\times\hat{\boldsymbol{x}}.
\end{equation}

Projecting into this plane gives load coordinates $(m_\ell,z_\ell)$ and a desired tangent angle 
\begin{align}
\boldsymbol{d}_{\mathrm{proj}} &=
  \boldsymbol{d}_{\mathrm{world}}
  -\hat{\boldsymbol{z}}\,(\hat{\boldsymbol{z}}^\top\boldsymbol{d}_{\mathrm{world}}), \\[6pt]
\phi_{\text{des}} &=
  \arctan\!\left(
    \frac{\boldsymbol{d}_{\text{proj}} \cdot \hat{\boldsymbol{y}}}
         {\boldsymbol{d}_{\text{proj}} \cdot \hat{\boldsymbol{x}}}
  \right).
\end{align}

\subsubsection{Outer--Inner Optimization Structure.}
The estimation problem is solved with a nested optimization design, where the outer loop optimizes the in-plane rotation $\alpha$, and the inner loop fits the catenary parameters $(a,c)$ for each candidate $\alpha$. The cable plane may be rotated to better align with the measured tangent. The rotation of the in-plane basis is
\begin{equation}
\begin{bmatrix}
\hat{\boldsymbol{x}}_\alpha \\[4pt]
\hat{\boldsymbol{y}}_\alpha
\end{bmatrix}
=
\begin{bmatrix}
\cos\alpha & \sin\alpha \\[4pt]
-\sin\alpha & \cos\alpha
\end{bmatrix}
\begin{bmatrix}
\hat{\boldsymbol{x}} \\[4pt]
\hat{\boldsymbol{y}}
\end{bmatrix},
\end{equation}
which induces rotated load coordinates
\begin{equation}
\begin{bmatrix}
m_\ell^\alpha \\[4pt]
z_\ell^\alpha
\end{bmatrix}
=
\begin{bmatrix}
\cos\alpha & \sin\alpha \\[4pt]
-\sin\alpha & \cos\alpha
\end{bmatrix}
\begin{bmatrix}
m_\ell \\[4pt]
z_\ell
\end{bmatrix}.
\end{equation}
At the load point, the predicted tangent angle is
\begin{equation}
\theta_{\text{model}}(\alpha) = \alpha + \arctan\!\left(\sinh\!\Big(\tfrac{m_\ell^\alpha - c}{a}\Big)\right),
\end{equation}
and the outer cost penalizes deviation from the desired tangent angle $\phi_{\text{des}}$, with a notch term to avoid degeneracy:
\begin{equation}
J_\theta(\alpha) = (\theta_{\text{model}}(\alpha) - \phi_{\text{des}})^2 
+ \beta \exp\!\Big(-\tfrac{(\alpha - \phi_{\text{des}})^2}{2\sigma^2}\Big),
\end{equation}
where the notch term is a Gaussian centered at $\phi_{\text{des}}$, with $\sigma$ controlling the spread and $\beta$ scaling the relative weight of this anti-degeneracy mechanism. Since $\alpha$ is unconstrained, we directly employ the L-BFGS method \cite{lbfgs} to optimize the outer loop.

For each candidate $\alpha$, the inner loop optimizes the catenary parameters $(a,c)$ by minimizing the weighted residuals of two physical constraints:
\begin{equation}
J(a,c,\alpha) = w_p\,\varepsilon_p^2 + w_L\,\varepsilon_L^2,
\end{equation}
with
\begin{align}
\varepsilon_p &= 
a\!\left[\cosh\!\Big(\tfrac{m_\ell^\alpha - c}{a}\Big) - \cosh\!\Big(\tfrac{c}{a}\Big)\right] - z_\ell^\alpha, \\[6pt]
\varepsilon_L &= 
a\sinh\!\Big(\tfrac{m_\ell^\alpha - c}{a}\Big) + a\sinh\!\Big(\tfrac{c}{a}\Big) - l_0,
\end{align}
where $\varepsilon_p$ enforces that the cable passes through the load and $\varepsilon_L$ ensures the arc length matches the cable length $l_0$. To eliminate explicit constraints on $(a,c)$ in this optimization, we employ a diffeomorphic reparameterization: we optimize over virtual variables $(v_a,v_c)\in\mathbb{R}^2$ and map to real parameters via
\begin{equation}
\begin{aligned}
a &= \begin{cases}
(0.5 v_a+1)v_a+1, & v_a>0,\\[3pt]
\dfrac{1}{(0.5 v_a-1)v_a+1}, & v_a\le0,
\end{cases} \\[6pt]
c &= m_\ell\!\left(\tfrac{\arctan v_c}{\pi}+\tfrac{1}{2}\right),
\end{aligned}
\end{equation}

which enforces $a>0$ and $c\in(0,m_\ell)$. The inverse maps (real$\!\to\!$virtual) are
\begin{equation}
\begin{aligned}
v_a &= 
\begin{cases}
\sqrt{2a-1}-1, & a>1,\\[4pt]
1-\sqrt{\tfrac{2}{a}-1}, & 0<a\le 1,
\end{cases} \\[6pt]
v_c &= \tan\!\Big(\pi\big(\tfrac{c}{m_\ell}-\tfrac{1}{2}\big)\Big),
\end{aligned}
\end{equation}

and gradients transform accordingly by the chain rule:
\begin{equation}
\begin{aligned}
\frac{\partial J}{\partial v_a} &= \frac{\partial J}{\partial a}\,\frac{\partial a}{\partial v_a}, \qquad
\frac{\partial J}{\partial v_c} = \frac{\partial J}{\partial c}\,\frac{\partial c}{\partial v_c}, \\[6pt]
\frac{\partial a}{\partial v_a} &= 
\begin{cases}
v_a+1, & v_a>0,\\[6pt]
\dfrac{1-v_a}{\big((0.5v_a-1)v_a+1\big)^2}, & v_a\le 0,
\end{cases} \\[6pt]
\frac{\partial c}{\partial v_c} &= \frac{m_\ell}{\pi(1+v_c^2)},
\end{aligned}
\end{equation}

with the reverse relation
\begin{equation}
\frac{\partial J}{\partial a}=\frac{\partial J}{\partial v_a}\,\Big/\frac{\partial a}{\partial v_a},\qquad
\frac{\partial J}{\partial c}=\frac{\partial J}{\partial v_c}\,\Big/\frac{\partial c}{\partial v_c}.
\end{equation}

After the diffeomorphic reparameterization, we optimize the inner variables $v_a$ and $v_c$ using the L-BFGS  method \cite{lbfgs}, and then apply the inverse diffeomorphism to recover the real parameters $a$ and $c$.

\subsubsection{Back to 3D.}
After optimizing $(a,c,\alpha)$, the cable curve in world coordinates is reconstructed as
\begin{equation}
\boldsymbol{r}(x) = \boldsymbol{p}_u + \hat{\boldsymbol{x}}_\alpha x + \hat{\boldsymbol{y}}_\alpha z(x), 
\quad x\in[0,m_\ell^\alpha],
\end{equation}
with
\begin{equation}
z(x) = a\Big[\cosh\!\Big(\tfrac{x-c}{a}\Big) - \cosh\!\Big(\tfrac{c}{a}\Big)\Big].
\end{equation}

\subsection{Cable Point-Cloud Filtering}

To suppress cable-induced artifacts in the LiDAR map, we build a volumetric mask around the cable centerline and remove points inside it before map integration.

\paragraph*{Taut-cable case.}
When the cable is taut, its centerline reduces to a straight segment connecting the UAV position $\boldsymbol{p}_Q$ and the payload position $\boldsymbol{p}_L$:
\[
\boldsymbol{r}_{\text{taut}}(x)
   = \boldsymbol{p}_Q
     + \frac{x}{\|\boldsymbol{p}_L-\boldsymbol{p}_Q\|}\,
       (\boldsymbol{p}_L-\boldsymbol{p}_Q),
   \quad x\in[0,l_0].
\]
Curve fitting is unnecessary; we uniformly sample this segment with spacing
\[
\Delta s = \frac{l_0}{N_s-1},
\]
and build spherical exclusion regions along it.

\paragraph*{bent-cable case.}
When the cable bends, we use the estimated curve $\boldsymbol{r}(x)$ in world coordinates as a geometric prior and sample it in the same way:
\[
\mathcal{P}_c = \{\, \boldsymbol{r}(x_i) \mid x_i = i\,\Delta s,\ i=0,\dots,N_s-1 \}.
\]
For each sample $\boldsymbol{r}(x_i)$, a spherical exclusion region is defined:
\[
\mathcal{B}_i
   = \big\{ \boldsymbol{p}\in\mathbb{R}^3
            \mid
            \| \boldsymbol{p}-\boldsymbol{r}(x_i)\|_2
              \le r_c \big\},
\]
where $r_c$ covers the cable radius and a noise margin.
At the terminal point (coincident with the payload), the radius is enlarged to $r_L$:
\[
\mathcal{B}_{\text{load}}
   = \big\{ \boldsymbol{p}\in\mathbb{R}^3
             \mid
             \| \boldsymbol{p}-\boldsymbol{r}(l_0)\|_2
                \le r_L \big\}.
\]

A sufficient condition for gap-free coverage is
\[
r_c \ge \frac{\Delta s}{2},
\qquad
r_L \ge \frac{\Delta s}{2},
\]
since the Euclidean distance between neighboring samples satisfies $d_i \le \Delta s$.
In practice we choose
\[
r_c \ge \max\big(r_{\text{cable}},\; \frac{\Delta s}{2}+\epsilon\big),
\]
where $r_{\text{cable}}$ is the physical cable radius and $\epsilon$ a sensor-noise margin.

Finally, the union
\[
\mathcal{B}
   = \bigcup_{i=0}^{N_s-1}\mathcal{B}_i
     \cup \mathcal{B}_{\text{load}}
\]
is used as a volumetric mask; LiDAR points ${}^{\mathcal{W}}\boldsymbol{p}_{Lj}$ lying in $\mathcal{B}$ are discarded, removing cable artifacts while retaining true structures and improving SLAM robustness.

\section{Corridor-Based Planning for Suspended-Load UAV Transport Systems}
\subsection{Safe Flight Corridor Generation for Multi-Size Aerial Transport}
\label{sec:maciri}
Motivated by the limitations of prior convex decomposition methods, we develop the \textbf{M}ulti-size-\textbf{A}ware \textbf{C}onfiguration-space \textbf{I}terative \textbf{R}egional \textbf{I}nflation (MACIRI) framework, building upon the iterative process of  CIRI \cite{super}. MACIRI generalizes corridor generation from point seeds to arbitrary convex-hull seeds and adapts to suspended-load UAV transport systems with varying size ratios. This enables the expansion of corridor volumes while guaranteeing 100\% containment of seeds for multi-size robotic systems.

The MACIRI procedure, as outlined in Algorithm~\ref{alg:MACIRI}, initiates with the ellipsoid as a unit ball (line~1) and proceeds through iterative steps until the ellipsoid's volume converges to a stable value (line~23). Each iteration consists of two main stages. In the first stage, the ellipsoid is inflated uniformly until it tangentially contacts the obstacle point cloud, leading to the generation of hyperplanes that partition a convex region from the obstacles, all while ensuring the containment of the seed (line~5-21). In the second stage, the SOCP-based optimization approach \cite{firi} is applied to determine the maximum-volume ellipsoid that can be inscribed within the polyhedron formed by the obstacle-free half-spaces defined by the previously generated hyperplanes (line~22). Through iterative refinement, both the ellipsoid and the hyperplanes are progressively adjusted, resulting in an expanded inscribed ellipsoid and a larger polyhedron that encompasses an increasingly larger obstacle-free space.

In the first stage of the iterative process, obstacle points are modeled as spheres whose radii are adaptively determined by the size of their nearest seed vertex (line~8-9). Subsequently, the Ellipsoid-based Convex Decomposition method (EllipsoidCvxDecomp) of CIRI \cite{super} is directly applied to compute separating hyperplanes between the ellipsoid and obstacle spheres (line~10). This adaptive treatment of vertex sizes maximizes usable free space without conservatively relying on the maximum size. Unlike single-size formulations, the proposed multi-size approach naturally adapts to diverse geometries of suspended-load UAV transport systems.  

However, EllipsoidCvxDecomp \cite{super} ensures containment only for point seeds, and CIRI \cite{super} only supports containment correction for line seeds. To overcome this limitation, we introduce a containment correction strategy that guarantees strict enclosure of all seed vertices. For convex-hull seeds, each face is decomposed into triangular facets without loss of generality. Specifically, the three nearest vertices of the convex-hull seed to the obstacle point cloud are identified, forming a triangular face (line~13). Then, the nearest point on this triangular face to the center of an obstacle sphere is computed (line~14), and the corresponding difference vector is used as the face normal when constructing tangent planes of the obstacle spheres (line~15-16). For line-seed corridors, the nearest point on the line segment is used analogously(line~18-20). This procedure ensures that the resulting corridor strictly contains the seeds.

The proposed MACIRI naturally extend to UAVs with suspended payloads, where seeds may take the form of line segments, quadrilaterals, or tetrahedra. The multi-size representation enables corridors that respect different UAV–payload size ratios, while the containment correction ensures that all geometric configurations are robustly enclosed. Together, these properties enable MACIRI to generate reliable safe flight corridors with larger volumes, ensuring 100\% containment, and tailored to suspended-load aerial transport systems.

\SetKwFor{For}{for}{\string do}{}
\RestyleAlgo{ruled}
\begin{algorithm}[h]
    \caption{MACIRI}
    \label{alg:MACIRI}
    \LinesNumbered
    \SetKwInOut{Notation}{Notation}

    \Notation{Identity $\boldsymbol{I}\!\in\!\mathbb{R}^{3\times3}$\; 
    Ellipsoid $\mathcal{E}$ centered at $\boldsymbol{d}$ with shape $\boldsymbol{C}$\; 
    The volume of $\mathcal{E}$ : $vol(\mathcal{E})$\;
    Convergence threshold $\rho$\;
    Active obstacle point cloud $\mathcal{O}_{a}$\; 
    Hyperplane $\mathcal{H}$; face normal $\boldsymbol{n}_f$\;
    Vertices of triangular faces $\mathcal{V}_{tri}$.}

    \KwIn{
    \begin{itemize}
        \item Obstacle point cloud $\mathcal{O}=\{\boldsymbol{o}_j\}$
        \item The $n$ vertices of the seed $\mathcal{V}=\{\boldsymbol{v}_j\}$ \\(for point, line, or convex-hull seeds)
        \item Vertex–radius map $\mathcal{R}_{\mathcal{V}}$
    \end{itemize}}
    \KwOut{Polyhedron $\mathcal{P}$}

    \SetKwFunction{MACIRI}{MACIRI}
    \SetKwProg{Fn}{Function}{:}{End Function}
    \Fn{\MACIRI{$\mathcal{O},\mathcal{V},\mathcal{R}_{\mathcal{V}}$}}{

        $\mathcal{E}.\boldsymbol{C} \leftarrow \boldsymbol{I}$\;
        $\mathcal{E}.\boldsymbol{d} \leftarrow \dfrac{1}{n}\sum_{\boldsymbol{v}_j \in \mathcal{V}} \boldsymbol{v}_j$\;
        \texttt{converge} $\leftarrow$ \textbf{false}\;

        \While{\textbf{not} \texttt{converge}}
        {
            $\mathcal{E}_{\text{last}} \leftarrow \mathcal{E}$\;

            \For{$\boldsymbol{o}_j \in \mathcal{O}$}{
                $\boldsymbol{v}^\star \leftarrow \textbf{NearestVertex}(\mathcal{V}, \boldsymbol{o}_j)$\;
                $r_j \leftarrow \mathcal{R}_{\mathcal{V}}(\boldsymbol{v}^\star)$\;
                % Inflate via convex decomposition in C-space
                $\mathcal{H} \leftarrow \textbf{EllipsoidCvxDecomp}(\boldsymbol{v}^\star, \boldsymbol{o}_j, r_j)$\;
                            % --- Containment correction (branch by seed type) ---

            \uIf{\textbf{not} $\textbf{IsSeedInPlane}(\mathcal{V}, \mathcal{H})$}
            {
                \uIf{$\textbf{IsConvexHull}(\mathcal{V})$}{
                $\mathcal{V}_{tri} \leftarrow \textbf{NearestTriangle}(\mathcal{V}, \boldsymbol{o}_j)$\;
                $\boldsymbol{p}^\star \leftarrow \textbf{NearestPoint}(\mathcal{V}_{tri}, \boldsymbol{o}_j)$\;
                $\boldsymbol{n}_f \leftarrow \boldsymbol{p}^\star - \boldsymbol{o}_j$\;
                $\mathcal{H} \leftarrow \textbf{GetTangentPlane}(\boldsymbol{o}_j, r_j, \boldsymbol{n}_f)$\;
                }
               \uElseIf{$\textbf{IsLine}(\mathcal{V})$}{
                    $\boldsymbol{p}^\star \leftarrow \textbf{NearestPoint}(\mathcal{V},\boldsymbol{o}_j)$\;
                    $\boldsymbol{n}_f \leftarrow \boldsymbol{p}^\star - \boldsymbol{o}_j$\;
                    $\mathcal{H} \leftarrow \textbf{GetTangentPlane}(\boldsymbol{o}_j, r_j, \boldsymbol{n}_f)$\;
            }
            }

            $\mathcal{P} \leftarrow \textbf{AddPlane}(\mathcal{P},\, \mathcal{H})$\;

            }

            % Update maximum-volume inscribed ellipsoid for next iteration
            $\mathcal{E} \leftarrow \textbf{MIVE}(\mathcal{P})$\;

            \uIf{$vol(\mathcal{E}) < (1 + \rho)vol(\mathcal{E}_{\text{last}})$}{
                \texttt{converge} $\leftarrow$ \textbf{true}\;
            }
        }

        \KwRet $\mathcal{P}$\;
    }
\end{algorithm}

\subsection{Kinodynamic A* as Seed Generation for MACIRI}
To generate the initial seeds for the MACIRI framework, we directly employ the Kinodynamic A* algorithm, which explicitly considers the dynamic constraints and potential collisions of both the UAV and the suspended load~\cite{autotrans}. This method simultaneously produces dynamically feasible and collision-free trajectories for both entities, which serve as high-quality initialization paths. These seed trajectories provide crucial guidance for the subsequent construction of the safe flight corridor within the MACIRI pipeline.

The Kinodynamic A* algorithm outputs two discrete trajectories:
\begin{equation}
\label{A*}
\pi_{\text{Q}} = \left\{ \boldsymbol{x}_{\text{Q}}^0, \boldsymbol{x}_{\text{Q}}^1, \dots, \boldsymbol{x}_{\text{Q}}^N \right\}, \quad
\pi_{\text{L}} = \left\{ \boldsymbol{x}_{\text{L}}^0, \boldsymbol{x}_{\text{L}}^1, \dots, \boldsymbol{x}_{\text{L}}^N \right\},
\end{equation}
where \( \pi_{\text{Q}} \) and \( \pi_{\text{L}} \) denote the position sequences of the UAV and the suspended load, respectively.

Based on the UAV trajectory \( \pi_{\text{Q}} \) and the load trajectory \( \pi_{\text{L}} \), we construct the seed set \( \boldsymbol{S} \) for initializing the MACIRI framework:
\begin{equation}
\label{A* seed}
\begin{aligned}
\boldsymbol{S} &= \left\{ \mathcal{V}_0, \mathcal{V}_1, \dots, \mathcal{V}_{N-1} \right\}, \\
\mathcal{V}_i &= \left\{ \boldsymbol{x}_{\text{Q}}^i, \boldsymbol{x}_{\text{Q}}^{i+1}, \boldsymbol{x}_{\text{L}}^i, \boldsymbol{x}_{\text{L}}^{i+1} \right\},
\end{aligned}
\end{equation}
where each \( \mathcal{V}_i \) is a set of vertices that can define either a quadrilateral or a tetrahedron, depending on the spatial configuration of the UAV and load states at adjacent time steps.

Subsequently, the seed set \( \boldsymbol{S} \) is provided as input to the MACIRI algorithm to generate the safe flight corridor \( \mathcal{F} \):
\begin{equation}
    \mathcal{F} = \left\{ \mathcal{P}_0, \mathcal{P}_1, \dots, \mathcal{P}_{N-1}, \dots, \mathcal{P}_{N+M-1} \right\},
\end{equation}
where \( N \) polytopes of  $\mathcal{F}$ are generated based on the seed set \( \boldsymbol{S} \), and the remaining \( M \) polytopes are constructed to ensure sufficient overlap between adjacent polytopes in \( \mathcal{F} \), thereby guaranteeing spatial continuity and feasibility for subsequent trajectory optimization.

\subsection{Trajectory Initialization for Suspended-Load Corridor Planning}
\label{sec:trans_init}
We propose a trajectory-initialization method for suspended-load corridor planning that reduces infeasibility in trajectory optimization. 
%------------------------------------------------------------------
\subsubsection{Problem Setup.}
Given a start $\boldsymbol{x}^{0}_{L}\in\mathbb{R}^3$, a goal $\boldsymbol{x}^{N+M}_{L}\in\mathbb{R}^3$, and a set $\mathcal{F}^*$ of overlapping polytopes $\mathcal{P}^{*}$ formed by adjacent polytopes $\mathcal{P}$
\begin{equation}
\begin{aligned}
\mathcal{F}^* &=  \{\mathcal{P}^{*}_1, \mathcal{P}^{*}_2, \dots,\mathcal{P}^{*}_{N+M-1} \}, \\
\mathcal{P}^{*}_i&= \mathcal{P}_{i-1} \cap \mathcal{P}_{i} 
\\&= \bigl\{\boldsymbol{x}\in\mathbb{R}^3\mid \boldsymbol{A}_i\boldsymbol{x}-\boldsymbol{b}_i\le\boldsymbol{0}\bigr\},
\end{aligned}
\end{equation}
we compute $N+M-1$ payload waypoints $\boldsymbol{x}^{i}_{L}\in\mathcal{P}^{*}_i$ and $N+M-1$ cable orientations $\boldsymbol{n}(\boldsymbol{u}_i)\in\mathbb{S}^2_+$ that jointly describe the coupled quadrotor–payload system.
Waypoints must stay inside $\mathcal{P}^{*}_i$, while cable orientations must satisfy $\|\boldsymbol{n}_i\|=1$ and $z_i>0$.

The initialization has three goals:
(i) generate a short initial path,
(ii) ensure the UAV and payload remain inside the safe corridor, and
(iii) enforce a separation margin between the cable midpoints and the corridor boundaries to enhance safety and robustness of the trajectory.
 
Accordingly, we construct the following objective:
\begin{align}
\label{eq:objective}
J=&\;
\underbrace{\sum_{i=0}^{N+M-1}\sqrt{\|\boldsymbol{x}^{i+1}_{L}-\boldsymbol{x}^{i}_{L}\|^{2}+\varepsilon}}_{\text{(1) short path}}
+\underbrace{\sum_{i=1}^{N+M-1}\phi(\boldsymbol{A}_{i}\boldsymbol{x}^{i}_{Q}-b_{i})}_{\text{(2) UAV in corridor}} \nonumber\\
&+\underbrace{\sum_{i=1}^{N+M-1}\operatorname{softmin}_{m}\!\bigl(-\boldsymbol{a}_{im}^{\!\top}\boldsymbol{x}^{i}_{mid}+d_{im}\bigr)}_{\text{(3) cable-midpoint separation margin}}.
\end{align}
Here $(\boldsymbol{a}_{im},d_{im})$ denote the half-spaces of $\mathcal{P}^{*}_i$, and $\varepsilon>0$ smooths the path-length term.  
For cell~$i$,
\begin{equation}
\begin{aligned}
\boldsymbol{x}^{i}_{Q}  &= \boldsymbol{x}^{i}_{L} + l_0\,\boldsymbol{n}_i, \\
\boldsymbol{x}^{i}_{mid}  &= \boldsymbol{x}^{i}_{L} + \tfrac{l_0}{2}\boldsymbol{n}_i,
\end{aligned}
\end{equation}
where $\boldsymbol{x}^{i}_{Q}$ denotes the UAV waypoint above payload waypoint $\boldsymbol{x}^{i}_{L}$ with cable length $l_0$, and $\boldsymbol{x}^{i}_{mid}$ is the midpoint of the cable connecting the UAV and the payload.  
And the function $\phi(\cdot)$ is a smooth penalty that grows rapidly for positive arguments and remains small when its argument is negative, thereby encouraging $\boldsymbol{A}_{i}\boldsymbol{x}^{i}_{Q}-b_{i}<0$.  
This ensures the UAV stays strictly inside each corridor polytope.

\subsubsection{Constraint Elimination via Diffeomorphic Reparameterization.}

We directly adopt the diffeomorphic reparameterization method proposed in~\cite{gcopter} to eliminate the polytope constraint on waypoints.  
Given a polytope $\mathcal{P}$ with vertices $\{\boldsymbol{v_0},\dots,\boldsymbol{v_{\hat{n}}\}}$, any point $\boldsymbol{x}_{L}\in\mathcal{P}$ can be represented in barycentric form as
\begin{equation}
\label{eq:BarycentricEuclideanTransformation}
\boldsymbol{x}_{L} = \boldsymbol{v_0} + \hat{\boldsymbol{V}} \boldsymbol{w}.
\end{equation}

Here $\hat{\boldsymbol{v}}_i = \boldsymbol{v}_i - \boldsymbol{v}_0$, 
$\hat{\boldsymbol{V}} = \{\hat{\boldsymbol{v}}_1,\dots,\hat{\boldsymbol{v}}_{\hat{n}}\}$, 
and $\boldsymbol{w}=\{w_1,\dots,w_{\hat{n}}\}^\top\in\mathbb{R}^{\hat{n}}$ is the last $\hat{n}$ entries in the barycentric coordinate, satisfying $\boldsymbol{w} \succeq \boldsymbol{0}$ and $\Vert{\boldsymbol{w}} \Vert_1 \leq 1$.  
The representation \cite{gcopter} can be reparameterized through a smooth surjection
$f_\mathcal{P}:\mathbb{R}^{\hat{n}}\to\mathcal{P}$ is
\begin{equation}
f_\mathcal{P}(\boldsymbol{\xi})=\boldsymbol{v_0}+\frac{4\hat{\boldsymbol{V}}[\boldsymbol{\xi}]^2}{(\boldsymbol{\xi}^\top \boldsymbol{\xi}+1)^2}\in\mathcal{P},\quad \forall \boldsymbol{\xi}\in\mathbb{R}^{\hat{n}}.
\end{equation}

Here $[\boldsymbol{\xi}]^2$ denotes the elementwise square.  
Thus any unconstrained variable $\boldsymbol{\xi}\in\mathbb{R}^{\hat{n}}$ is mapped to a feasible point in $\mathcal{P}$.  
Since $f_\mathcal{P}$ is differentiable, gradients can be propagated through the reparameterization.  
If $\boldsymbol{g}_i=\partial J/\partial \boldsymbol{x}^{i}_{L}$ is the $i$-th component of $\partial J/\partial\boldsymbol{x}_{L}$, then
\begin{equation}
\frac{\partial J}{\partial \xi_i}=\frac{8\xi_i\circ \hat{\boldsymbol{V}}^\top \boldsymbol{g}_i}{(\xi_i^\top\xi_i+1)^2}
-\frac{16\boldsymbol{g}_i^\top \hat{\boldsymbol{V}}[\xi_i]^2}{(\xi_i^\top\xi_i+1)^3}\xi_i.
\end{equation}
Thus, the polytope constraint is exactly removed via a smooth reparameterization, enabling unconstrained gradient-based optimization over waypoints.

Considering the constraint of the cable directions, we introduce a smooth bijection $F:\mathbb{R}^2\!\to\!\mathbb{S}^2_+$ from unconstrained virtual variables $\boldsymbol{u}_i=[u_i,v_i]^\top$ to the cable orientation $\boldsymbol{n}(\boldsymbol{u}_i)$:
\begin{equation}
\begin{aligned}
\boldsymbol{n}(\boldsymbol{u}_i) &= [x_i,y_i,z_i]^\top, \\[4pt]
x_i &= \frac{u_i}{\sqrt{1+u_i^{2}+v_i^{2}}}, \\[4pt]
y_i &= \frac{v_i}{\sqrt{1+u_i^{2}+v_i^{2}}}, \\[4pt]
z_i &= \sqrt{1-x_i^{2}-y_i^{2}}.
\end{aligned}
\label{eq:uv2xyz}
\end{equation}

This diffeomorphism ensures $\|\boldsymbol{n}_i\|=1$ and $z_i>0$ automatically, so cable feasibility is guaranteed by construction.

For a real 2D cable direction $\boldsymbol{r}=[x,y]^\top$ with $x^2+y^2<1$, the inverse of \eqref{eq:uv2xyz} is
\begin{equation}
\boldsymbol{u}=F^{-1}(\boldsymbol{r})=
\begin{bmatrix}
\dfrac{x}{\sqrt{1-x^{2}-y^{2}}}\\[6pt]
\dfrac{y}{\sqrt{1-x^{2}-y^{2}}}
\end{bmatrix}.
\end{equation}

Given the gradient of the real 2D cable direction $\partial J/\partial\boldsymbol{r}$, the chain rule yields the virtual-space gradient
\begin{equation}
\frac{\partial J}{\partial\boldsymbol{u}}=
\frac{1}{(1+x^{2}+y^{2})^{3/2}}
\begin{bmatrix}
1+y^{2} & -xy\\
-xy & 1+x^{2}
\end{bmatrix}\frac{\partial J}{\partial\boldsymbol{r}}.
\end{equation}

With both the waypoint and cable constraints eliminated by these diffeomorphic reparameterizations, the optimization problem becomes smooth and unconstrained in $\{\boldsymbol{\xi}_i,\boldsymbol{u}_i\}_{i=1}^{N+M-1}$.  
We minimize the objective~\eqref{eq:objective} using the L-BFGS algorithm \cite{lbfgs}, and all gradients are computed analytically via the above Jacobians to ensure efficient convergence.

\subsection{Suspended-Load Trajectory Generation Within Safe Flight Corridors}
In this section, we formulate suspended-load trajectory generation within safe flight corridors as a spatio-temporal optimization problem based on differential flatness (Section \ref{sec:flatness}).
\subsubsection{Trajectory Representation for Optimization}
We represent the trajectory using $M$ polynomial segments of dimension $D$, each of degree $N=2s-1$. A segment is expressed as
\[
\mathcal{Z}_{i}(t) = \boldsymbol{c}_{i}^{\top}\boldsymbol{\beta}(t), 
\quad \boldsymbol{\beta}(t) = \left[1, t_i, \ldots, t_i^{2s-1}\right]^{\top}.
\]
In this work, we set $D=3$ and $N=7$ to describe the payload position $\boldsymbol{x}_L$ in the flat-output space.

The optimization objective consists of control effort and a time-regularization term, as formulated in \eqref{equ:opt_cost}. Here, $Q_{J}$ weights the control energy, while $\lambda_{T}$ penalizes the total duration:
\begin{subequations}
\begin{align}
\min_{\boldsymbol{c},\boldsymbol{T}} \; J(\boldsymbol{c},\boldsymbol{T}) 
&= \underbrace{\sum_{i=1}^{M}\int_{0}^{t_i} 
\mathcal{Z}_{i}^{(s)}(t)\, Q_{J}\, \mathcal{Z}_{i}^{(s)}(t)\,dt}_{\text{Energy Cost}}
+ \underbrace{\lambda_{T}\,\|\boldsymbol{T}\|}_{\text{Time Cost}}, \label{equ:opt_cost}\\
\text{s.t.}\quad 
& \mathcal{Z}_{1}^{[s-1]}(0) = \boldsymbol{z}_{0}, \quad 
\mathcal{Z}_{M}^{[s-1]}(t_{M}) = \boldsymbol{z}_{f}, \label{equ:opt_boundary_conditions}\\
& \mathcal{Z}_{i}^{[s-1]}(t_{i}) = \mathcal{Z}_{i+1}^{[s-1]}(0), \label{equ:opt_continuous_constraint}\\
& t_i > 0, \quad t_i \in \boldsymbol{T}, \label{equ:time_positive}\\
& \mathcal{C}_{b}\!\left(\mathcal{Z}_{i}(t), \mathcal{Z}_{i}^{(1)}(t), \ldots, \mathcal{Z}_{i}^{(s)}(t)\right) \leq 0, \quad b \in \mathcal{G}. \label{equ:constraints}
\end{align}
\end{subequations}
In this formulation, $\mathcal{C}_{b}$ represents violation functions enforcing dynamics and collision-avoidance constraints, and $\mathcal{G}$ denotes the complete constraint set. Equations \eqref{equ:opt_boundary_conditions} and \eqref{equ:opt_continuous_constraint} impose boundary and continuity conditions, respectively. Here, $\mathcal{Z}_{i}^{(s)}$ indicates the $s$-th derivative of the flat output $\mathcal{Z}_{i}$, while $\mathcal{Z}_{i}^{[s-1]}$ denotes all derivatives from order $0$ up to $s-1$.

To deform the spatio-temporal trajectory, we adopt the MINCO representation \cite{gcopter}, which allows simultaneous optimization of both time allocation and trajectory shape, while inherently satisfying boundary \eqref{equ:opt_boundary_conditions} and continuity \eqref{equ:opt_continuous_constraint} constraints. Instead of optimizing directly over the coefficient vectors $\boldsymbol{c}$, MINCO parameterizes the trajectory by the initial and terminal states, a set of intermediate waypoints $\boldsymbol{p}$ (Section~\ref{sec:trans_init}), and the segment durations $\boldsymbol{T}$. The trajectory is guaranteed to pass through $\boldsymbol{p}$ within the specified $\boldsymbol{T}$. The transformation 
$\boldsymbol{c} = \mathcal{M}(\boldsymbol{p}, \boldsymbol{T})$,
reduces the number of optimization variables and improves numerical stability. Moreover, the mapping $\mathcal{M}$ is smooth, computationally efficient with linear complexity, and differentiable. In this work, the segment durations $\boldsymbol{T}$ is initialized to the distance between adjacent waypoints $\boldsymbol{p}$ divided by the maximum speed $v_{\max}$.

To handle continuous-time constraints $\mathcal{C}_{b}$, we adopt a time-integral penalty formulation as proposed in~\cite{gcopter}. Instead of enforcing $\mathcal{C}_{b}$ at all time instances—which would result in infinitely many constraints—the violation is reformulated as a continuous-time functional:
\begin{equation}
\label{eq:ConstraintTranscription}
I_{\mathcal{C}_{b}}^k[p] = \int_0^T \max\left( \mathcal{C}_{b}(p(t), \dots, p^{(s)}(t)), \boldsymbol{0} \right)^k dt,
\end{equation}
where $k > 1$ ensures differentiability. A weighted version of this functional is defined as:
\begin{equation}
I_{\mathcal{C}_{b}}[p] = \chi^\top I_{\mathcal{C}_{b}}^k[p],
\end{equation}
where $\chi \in \mathbb{R}^{n_g}_{\geq 0}$ is a vector of weights reflecting the importance of each constraint. This penalty remains zero when constraints are satisfied and increases rapidly in the presence of violations.

For practical implementation, we follow the numerical quadrature method described in~\cite{gcopter}, which samples the trajectory at normalized timestamps and uses the trapezoidal weights rule to approximate the integral. This results in a differentiable and scalable approximation of the constraint violation functional. The integrand is evaluated at discrete points within each segment and aggregated using a weighted sum:
\begin{equation}
\label{eq:TimeIntegralPenalty}
I(\boldsymbol{c}, \boldsymbol{T}) = \sum_{i=1}^{M} \frac{T_i}{\kappa_i} \sum_{j=0}^{\kappa_i} \bar{\omega}_j\, \chi^\top \max\left( \mathcal{C}_{b\tau}\left(\boldsymbol{c}_i, T_i, \frac{j}{\kappa_i} \right), \boldsymbol{0} \right)^k,
\end{equation}
where $\mathcal{C}_{b\tau}(\boldsymbol{c}_i, T_i, \tau)$ denotes the sampled constraint function at normalized time $\tau \in [0,1]$ for segment $i$, $\kappa_i$ is the number of sampling points, and $\bar{\omega}_j$ are the trapezoidal weights, with $\bar{\omega}_0 = \bar{\omega}_{\kappa_i} = 1/2$ and $\bar{\omega}_j = 1$ otherwise. This formulation enables parallel evaluation and smooth gradient computation, making it well-suited for optimization-based trajectory planning.

\subsubsection{Corridor-based Whole-body Safety Guarantee}
\label{sec:corridor_based_collision}
Rather than explicitly sampling along the cable to approximate the suspended-load system \cite{autotrans}, we exploit the inclusion property of the safe flight corridors (Section~\ref{sec:property_corr}). In this formulation, it is sufficient to constrain only the quadrotor and payload positions, while the inclusion property ensures that the entire system remains collision-free. This avoids cable sampling and significantly reduces optimization time.  

Formally, the safe corridor constraint is described as a linear inequality constraint,
\begin{equation}
    \label{equ:corridor_constraint}
    \mathcal{C}_{bcorr} = 
    \begin{cases}
    \boldsymbol{A}_i\boldsymbol{x}_Q \leq b_i ,\\
    \boldsymbol{A}_i\boldsymbol{x}_L \leq b_i,
    \end{cases}
\end{equation}
with $\boldsymbol{A}_i\in\mathbb{R}^{m\times 3}$ and $b_i\in\mathbb{R}^m$ defining the corridor facets. It guarantees that both the quadrotor position $\boldsymbol{x}_Q$ and payload position $\boldsymbol{x}_L$ remain inside the corridor. And  $\boldsymbol{x}_Q$ can be obtained by Equation ~\eqref{eq:taut_geom} and Equation ~\eqref{equ:flat_rot_ext_b}.

In contrast to ESDF-based MINCO approaches~\cite{autotrans, impactor}, which enforce obstacle-avoidance constraints at discretely sampled points along the trajectory and therefore cannot guarantee safety for the unsampled states in between, our corridor-based method offers a significantly stronger probabilistic safety assurance across the entire continuous trajectory. 

We define a suspended-load seed as a pair comprising the positions of the quadrotor and the payload. For any two consecutive seeds lying within the same corridor segment, we construct a convex region that is either a quadrilateral or a tetrahedron. Ensuring that both seeds are collision-free implies that the convex hull formed between them is also free of obstacles, as discussed in Section~\ref{sec:property_corr}. 

By sequentially linking such safe convex regions, we obtain a continuous chain of $n$ connected convex hull. This chain is highly likely to contain the unsampled trajectory states, thereby mitigating the risks of inter-sample collisions inherent in ESDF-based methods.

\subsubsection{Dynamic Constraints} 
To guarantee the feasibility of the generated trajectory, we enforce dynamic constraints derived from the differential flatness property of the system.

\subsubsection*{Quadrotor actuator constraints}
The quadrotor is subject to hardware limitations, in particular bounds on thrust and maximum tilt. From the flatness formulation in Section~\ref{sec:flatness}, the  thrust $f$ and the thrust direction ${}^{\mathcal{W}}{R_{\mathcal{Q}}}\boldsymbol{e}_{z}$ can be obtained by Equation~\eqref{equ:flat_rot_ext_c}.
The thrust magnitude is restricted to remain within $[f_{l},f_{u}]$ through
\begin{equation}
    \label{equ:thrustconstraint}
    \mathcal{C}_{bf}(\boldsymbol{z}) = \left(f-\tfrac{f_{u}+f_{l}}{2}\right)^{2} - \left(\tfrac{f_{u}-f_{l}}{2}\right)^{2} \leq 0,
\end{equation} 
which stabilizes the vehicle’s attitude and avoids dynamically infeasible solutions.  
In addition, the tilt angle of the quadrotor is bounded by $\theta_{\max}$ to prevent loss of feasibility and flipping. This requirement is encoded as
\begin{equation}
    \label{equ:tiltangleconstraint}
    \mathcal{C}_{b\theta}(\boldsymbol{z}) = \cos(\theta_{\max}) - \boldsymbol{e}_{z}^{\top}{}^{\mathcal{W}}{R_{\mathcal{Q}}}\boldsymbol{e}_{z} \leq 0.
\end{equation} 

\subsubsection*{Payload dynamics constraints}
To avoid excessive motion of the suspended payload, its velocity and acceleration are constrained as
\begin{equation}
    \label{equ:payloaddynamicsconstraint}
    \mathcal{C}_{bp}(\boldsymbol{z}) =
    \begin{cases}
        \|\dot{\boldsymbol{x}}_{L}\|_{2}^{2} - v_{\max}^{2} \leq 0, \\[4pt]
        \|\ddot{\boldsymbol{x}}_{L}\|_{2}^{2} - a_{\max}^{2} \leq 0.
    \end{cases}
\end{equation} 

\subsubsection*{Cable tension constraint}
As stated in Section~\ref{sec:nominalsystemdynamics}, the cable is assumed taut at all times. From Equation~\eqref{equ:flat_rho_ext_a} and Equation~\eqref{equ:flat_rot_ext_b}, the cable tension is given by  
$f_{T} = \| m_{L}(\ddot{\boldsymbol{x}}_{L} + g\boldsymbol{e}_{z})-\boldsymbol{F_L}\|_{2}$.  
Projecting onto the $z$-axis, the non-bent condition is expressed as
\begin{equation}
    \label{equ:cabletensionconstraint}
    \mathcal{C}_{bf_T}(\boldsymbol{z}) = \epsilon - \ddot{\boldsymbol{x}}_{L}\boldsymbol{e}_{z} - g \leq 0,
\end{equation} 
where $\epsilon$ is a small positive margin introduced to mitigate numerical errors. 

\subsubsection{Unconstrained Optimization}
To accelerate computation, all constraints are absorbed into a penalty formulation, transforming the problem into an unconstrained one. Since each segment duration must remain strictly positive (see \eqref{equ:time_positive}), we introduce a set of virtual times $\boldsymbol{\sigma}=[\sigma_1,\ldots,\sigma_n]\in\mathbb{R}^n$ and apply a diffeomorphic mapping \cite{gcopter} to obtain the true durations $t_i\in\mathbb{R}^+$.  

Constraint violations are accumulated using an integral penalty,
\begin{equation}
    \label{equ:discreteintegration}
    S = \sum_{i=1}^{M} \int_{0}^{t_{i}} L_{1}(\mathcal{C}_{b}(\boldsymbol{z})) \, dt,
\end{equation}
where $L_{1}(\cdot)$ denotes a smooth penalty function \cite{smoothL1}.  
The overall trajectory optimization problem is therefore reformulated as
\begin{equation}
    \label{equ:unconopt}
    \min_{\boldsymbol{q},\boldsymbol{\sigma}} \; J(\boldsymbol{q},\boldsymbol{\sigma}) 
    = \sum_{i=1}^{M}\int_{0}^{t_{i}} \mathcal{Z}_{i}^{(s)}(t) Q_{J}\mathcal{Z}_{i}^{(s)}(t) \, dt
    + \lambda_{T}\|\boldsymbol{T}\| + \lambda_{s} S,
\end{equation}
with $\lambda_{s}$ weighting the contribution of constraint penalties.

\section{Corridor Constraints for Nonlinear Model Predictive Control with Cable Bending}
\label{sec:nmpc_bend}
\subsection{Hybrid Dynamic Model for Suspended-Load UAV Systems}
In the hybrid dynamic framework (Section~\ref{sec:nominalsystemdynamics}), the payload system states are expressed as
\begin{equation}
x = \begin{bmatrix} \boldsymbol{x}_L,\, \boldsymbol{x}_Q,\, \dot{\boldsymbol{x}}_L,\, \dot{\boldsymbol{x}}_Q,\, {}^{\mathcal{W}}{R_{\mathcal{Q}}},\, \boldsymbol{\rho},\, \dot{\boldsymbol{\rho}} \end{bmatrix}^\top,
\end{equation}
which collectively describe the payload and quadrotor positions and velocities, 
the quadrotor's attitude, as well as the cable orientation and its time derivative. 
When the switch variable $s$ is set to $s=1$, the system operates in the taut-cable mode, whereas $s=0$ indicates that the cable is bent.
The control input vector is defined as
\begin{equation}
\boldsymbol{u} = \begin{bmatrix} f,\, \boldsymbol{\omega}_Q \end{bmatrix}^\top,
\end{equation}
By applying the Lagrange--d'Alembert formulation, the differential equations of motion 
for the coupled system are obtained:

\begin{subequations}
\begin{align}
\label{eq:diff_eq}
\dot{\boldsymbol{x}}_Q &= \boldsymbol{v}_Q  \\[6pt]
\ddot{\boldsymbol{x}}_Q &= s f_T \boldsymbol{\rho} + \frac{f {}^{\mathcal{W}}{R_{\mathcal{Q}}} \boldsymbol{e}_z + \boldsymbol{F}_Q}{m_Q} - g \boldsymbol{e}_z  \\[6pt]
\dot{\boldsymbol{x}}_L &= v_L = \frac{d\boldsymbol{x}_L}{dt}  \\[6pt]
\ddot{\boldsymbol{x}}_L &= - s f_T \boldsymbol{\rho} + \frac{\boldsymbol{F}_L}{m_L} - g \boldsymbol{e}_z  \\[6pt]
\dot{\boldsymbol{\rho}} &= \frac{d\boldsymbol{\rho}}{dt}  \\[6pt]
\ddot{\boldsymbol{\rho}} &= s \Bigg( \boldsymbol{\rho} \times \Bigg( \boldsymbol{\rho} \times 
      \left( \frac{f {}^{\mathcal{W}}{R_{\mathcal{Q}}} \boldsymbol{e}_z + \boldsymbol{F}_Q}{l m_Q} - \frac{\boldsymbol{F}_L}{l m_L} \right) \Bigg) \Bigg) \notag \\ 
      &\quad - s (\dot{\boldsymbol{\rho}} \cdot \dot{\boldsymbol{\rho}}) \boldsymbol{\rho} 
      - (1-s)\left( -\frac{\boldsymbol{F}_L}{m_L} + \frac{f {}^{\mathcal{W}}{R_{\mathcal{Q}}} \boldsymbol{e}_z +  \boldsymbol{F}_Q}{m_Q} \right)  \\[6pt]
 {}^{\mathcal{W}}{\dot{R}_{\mathcal{Q}}} &= {}^{\mathcal{W}}{R_{\mathcal{Q}}} \hat{\omega}_Q  \\[6pt]
f_T &= \frac{\boldsymbol{\rho} \cdot \big( m_Q + \boldsymbol{F}_L - m_L (f {}^{\mathcal{W}}{R_{\mathcal{Q}}} \boldsymbol{e}_z + \boldsymbol{F}_Q)\big)}{m_Q(m_Q+m_L)}  \notag \\ &\quad
      + \frac{m_L l (\dot{\boldsymbol{\rho}} \cdot \dot{\boldsymbol{\rho}})}{(m_Q+m_L)} 
\end{align}
\end{subequations}

\subsection{Model Predictive Control Formulation with Cable-Bending}
From the bent cable shape estimation derived earlier (Section~\ref{sec:cable_shape_etimator}), we construct a conservative trapezoidal envelope \(\mathcal{T}\) at each prediction step \(k\).  
The trapezoid is determined from endpoint tangents, the direct endpoint connection, and the tangent at the lowest bending point, resulting in four vertices \(\{\boldsymbol{v}_{j}\}_{j=1}^{4}\) with
\begin{equation}
\mathcal{T}=\operatorname{conv}\{\boldsymbol{v}_{1},\boldsymbol{v}_{2},\boldsymbol{v}_{3},\boldsymbol{v}_{4}\}.
\end{equation}
This trapezoid $\mathcal{T}$ fully encloses the UAV, payload, and bent cable, while remaining relatively compact so as to preserve optimization freedom:
\begin{equation}
\mathcal{B} \;\subseteq\; \mathcal{T} ,
\end{equation}
where \(\mathcal{B}\) denotes the set of all UAV, payload, and bent cable positions.

Rather than constraining the continuous cable curve, we then impose constraints only on the trapezoid vertices:
\begin{equation}
A\,\boldsymbol{v}_{j}\le b,\qquad j=1,\dots,4.
\end{equation}
And it guarantees \(\mathcal{T}\subseteq \mathcal{P}\).  
Since \(\mathcal{B}\subseteq\mathcal{T}\), all UAV, payload, and bent cable are safely contained without explicitly constraining every cable point, 
while the formulation requires only four vertex constraints (Section~\ref{sec:property_corr}), greatly reducing the number of inequalities in NMPC.

Finally, a quadratic optimization problem is formulated within the framework of NMPC, utilizing the hybrid dynamics model. The NMPC  problem is expressed as
\begin{subequations}
\label{eq:nmpc_cost}
\begin{align}
\min_{\boldsymbol{u}} \; \mathcal{J} &= 
\frac{1}{2}\sum_{k=1}^{H} 
\left( \|\boldsymbol{x}(k)-\boldsymbol{x}_r(k)\|_{Q_k}^2 + \|\boldsymbol{u}(k)-\boldsymbol{u}_r(k)\|_{R_k}^2 \right) \notag \\
& \quad + \frac{1}{2}\|\boldsymbol{x}(H)-\boldsymbol{x}_r(H)\|_{Q_H}^2 ,
\end{align}
subject to
\begin{align}
\label{eq:nmpc_dicrete_eq}
\boldsymbol{x}(k+1) &= f(\boldsymbol{x}(k),\boldsymbol{u}(k),s(k))  \\
\label{eq:nmpc_u}
\boldsymbol{u}_{\min} &\leq \boldsymbol{u} \leq \boldsymbol{u}_{\max},  \\
\label{eq:nmpc_bent_cable}
A\,\boldsymbol{v}_{j}&\le b,\qquad j=1,\dots,4, 
\end{align}
\end{subequations}
where $H$ denotes the prediction horizon, and $\boldsymbol{x}_r(k)$ and $\boldsymbol{u}_r(k)$ represent the reference state and input trajectory, respectively, at time step $k$, typically obtained from polynomial planning. In Equation~\eqref{eq:nmpc_dicrete_eq}, 
$f(\boldsymbol{x}(k),\boldsymbol{u}(k),s(k))$ is the discretized form of the hybrid dynamics, where $\boldsymbol{x}(k)$ and $\boldsymbol{u}(k)$ are the system states and control inputs, and $s(k)$ is the switching variable defined earlier. The input vector $\boldsymbol{u} = [f, \boldsymbol{\omega}_Q]^T$ is constrained as in Equation~\eqref{eq:nmpc_u} to ensure the thrust $f$ and angular rate $\boldsymbol{\omega}_Q$ of the quadrotor remain within feasible bounds. And the obstacle-avoidance with cable-bending is constrained as in Equation~\eqref{eq:nmpc_bent_cable}.

To enhance robustness against disturbances and estimation noise, exponential decay ~\cite{NMPC_QR_1, NMPC_QR_2} is applied to the weighting matrices $Q$ and $R$ as follows:
\begin{align}
Q_k &= \exp\!\left(-\tfrac{k}{H}b_x\right)Q, 
& R_k &= \exp\!\left(-\tfrac{k}{H}b_u\right)R, 
\end{align}
where $b_x$ and $b_u$ determine the decay rate.

For efficient real-time implementation, a warm-start strategy is employed to reduce computational effort. An initial guess is extracted from the reference trajectory provided by the planner through the differential flatness (Section~\ref{sec:flatness}). 

\section{Experiment And Evaluation}
\label{sec:experiment}
\subsection{System Setup}
\label{sec:system setup}
To experimentally validate the proposed method, we built a custom load-carrying unmanned aerial platform designed for full-body perception.

The platform is equipped with a PX4 flight control system, an Intel NUC 13 as the onboard computing unit, and two LiDAR sensors: an upward-facing LiDAR and a downward-facing LiDAR with a full hemispherical field of view. The upward-facing LiDAR is mounted at an inclined forward angle to enhance perception in the flight direction. A high-precision 9-axis IMU is installed at the end of the tether, while reflective strips are attached to the payload to facilitate LiDAR-based detection.

This design is particularly well-suited for load-carrying multirotor systems requiring full-body perception. Unlike conventional single-LiDAR UAVs, we integrated an additional downward-facing LiDAR. With its 360°×90° wide field of view, this sensor enables stable detection of the payload beneath the UAV during both daytime and nighttime, when vision sensors are prone to degradation. Moreover, it effectively perceives obstacles around the payload’s trajectory, which is critical for ensuring safe obstacle avoidance.

More importantly, the addition of the downward LiDAR significantly improves the UAV’s autonomous localization capability. Load-carrying UAVs frequently operate at high altitudes, such as above forests or between buildings. In such environments, localization relying solely on the upward LiDAR tends to degrade. By incorporating downward LiDAR point clouds, the system achieves much stronger environmental adaptability and robustness in localization.

In addition, the high-precision 9-axis IMU we selected provides more accurate attitude estimation. With its built-in high-precision magnetometer, it can effectively correct long-term drift in attitude during flight. By mounting this IMU at the end of the tether, high-frequency attitude information of the tether can be obtained, enabling precise modeling of its bending behavior.

\begin{figure}[!t]
\centering
\includegraphics[width=0.45\textwidth]{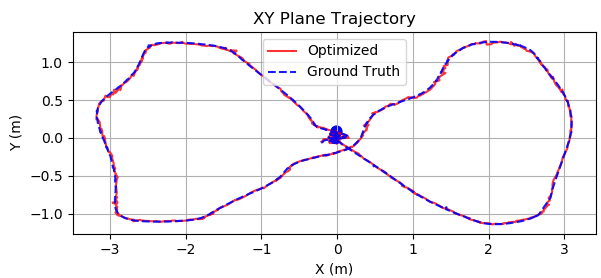}
\caption{Trajectory of the payload in the XY plane in estimation
 experiment.}
\label{fig:traj_xy}
\end{figure}

\begin{figure}[!t]
\centering
\includegraphics[width=0.45\textwidth]{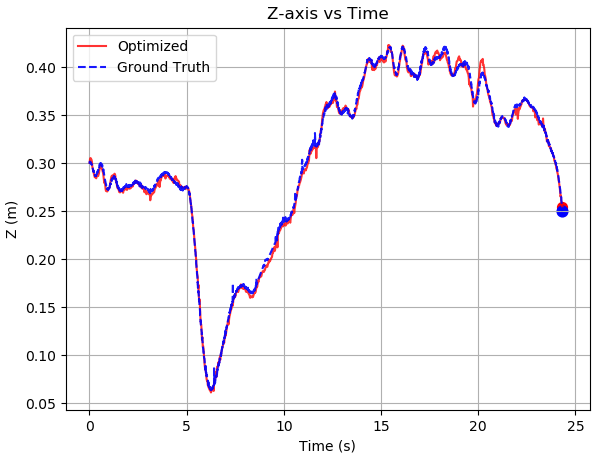}
\caption{Trajectory of the payload in the Z axis in estimation
 experiment.}
\label{fig:traj_z}
\end{figure}

\subsection{Performance of Whole-Body Perception}

\subsubsection{Pose Estimation of Suspended Loads}

To evaluate the accuracy of the proposed trajectory–estimation algorithm, the UAV was commanded to follow a figure-eight path at a nominal speed of 1.2 m/s—a manoeuvre that provides rich excitation in both translational and rotational degrees of freedom.The cable's length is 0.64 m. Figure \ref{fig:traj_xy} and \ref{fig:traj_z} compares the trajectory estimated by our method with the ground truth recorded by a NOKOV optical motion-capture system, showing a close agreement and thus confirming the effectiveness of the proposed approach.The specific estimation indicators are shown in the table \ref{tab:estimation indicator}.The estimation framework we designed can run over 100 Hz on our computing platform.

\begin{table}[tbp]
\centering
\caption{Estimation indicators}
\begin{tabular*}{0.8\linewidth}{@{\extracolsep{\fill}} l cc }
\hline
MAE [m] & RMSE [m] & Max [m] \\
\hline
0.015 & 0.018 & 0.078 \\
\hline
\end{tabular*}
\label{tab:estimation indicator}
\end{table}

\subsubsection{Disturbance-Aware Estimation of the Bent Cable Shape}
To evaluate the performance of bent-cable estimation under disturbances, ten different 3D wind fields were randomly generated within the range of $-4$ to $4~\mathrm{m/s}$ along each axis and applied to the cable. Three groups of bent cables with lengths of $0.64~\mathrm{m}$, $0.96~\mathrm{m}$, and $1.28~\mathrm{m}$ were considered for evaluation. For each cable, we obtained the ground-truth positions of $n$ sampled points from the MuJoCo simulator \cite{mujoco} and compared them against our estimated results. Specifically, $20$ samples were used for the $0.64~\mathrm{m}$ cable, $40$ samples for the $0.96~\mathrm{m}$ cable, and $60$ samples for the $1.28~\mathrm{m}$ cable. 

\begin{figure}[tbp]
\centering
\includegraphics[width=0.48\textwidth]{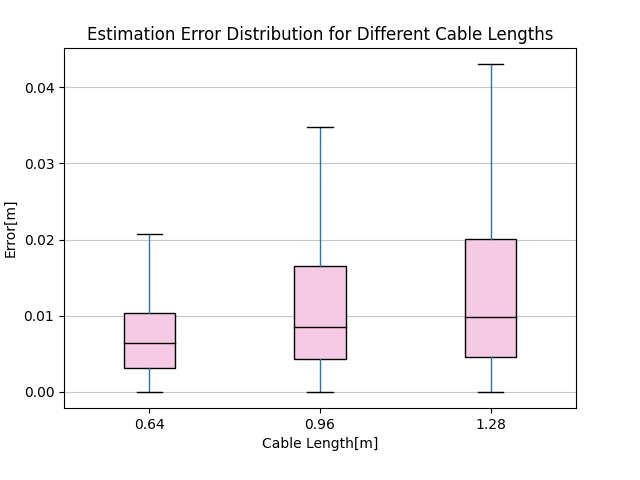}
\caption{Estimation error of bent-cable shapes under different wind disturbances.}
\label{fig:cable_error}
\end{figure}

The estimation errors are reported in Fig.~\ref{fig:cable_error}, and the computation times are summarized in Table~\ref{tab:cable_time}. The simulation results show that the majority of estimation errors remain below $2~\mathrm{cm}$, while the estimation frame rate of our method exceeds $5000~\mathrm{Hz}$, which is sufficient to meet the requirements of real-time bent-cable point cloud filtering and cable-obstacle avoidance control.

\begin{table}[h!]
\centering
\caption{Computation Time for Different Cable Lengths}
\begin{tabular*}{0.8\linewidth}{@{\extracolsep{\fill}} l cc }
\hline
Cable Length [m] & Avg [ms] & Max [ms] \\
\hline
0.64 & 0.1279 & 0.2023 \\
0.96 & 0.1453 & 0.2246 \\
1.28 & 0.1183 & 0.1548 \\
\hline
\end{tabular*}
\label{tab:cable_time}
\end{table}

\subsection{Performance of Generating Free Convex Polytope}
Building upon the analysis and comparison of several state-of-the-art (SOTA) algorithms for generating free convex polytopes presented in Section \ref{section:corridor_related_work}, we benchmark the proposed MACIRI algorithm against CIRI \cite{super}, FIRI-Shrinkage \cite{firi}, and FIRI-Inflation \cite{firi}. Notably, IRIS \cite{iris} and RILS \cite{rils} are excluded from this comparison, as they do not support convex-hull seeds as input. In order to systematically evaluate the corridor generation performance of the suspended-load UAV system under varying conditions, seven distinct size ratios were considered in the experiments. The size ratio, defined as the size of the payload divided by the size of UAV, took the following values: $20$, $10$, $5$, $1$, $\tfrac{1}{5}$, $\tfrac{1}{10}$, and $\tfrac{1}{20}$. We conduct the benchmark in a complex environment of  $50 \times 50 \times 5$ m size, where random obstacles are generated using Perlin noise \cite{Perlin}. For each test, a collision-free seed is randomly generated within the environment. The boundary for each convex hull generation algorithm is restricted to a  cube  with a side length of 2 m, centered at the seed's location and aligned with the coordinate axes. The obstacle input consists of points located within the square’s boundary on the map. To simulate the geometry of the suspended-load UAV system, we generate 100 random rectangular obstacles with  $1.0 \times 1.0 \times 0.64$ m size. For size ratios of $20$, $10$, $5$, and $1$, the size of the UAV is fixed at $0.02 \, \text{m}$, such that the maximum payload size can reach $0.4 \, \text{m}$. Conversely, for size ratios of $\tfrac{1}{5}$, $\tfrac{1}{10}$, and $\tfrac{1}{20}$, the size of the payload is fixed at $0.02 \, \text{m}$, allowing the UAV to achieve a maximum size of $0.4 \, \text{m}$. To ensure the solvability of the problem, we constrain the randomly generated seed such that the distance between the seed and any obstacle is greater than $0.4 \, \text{m}$.

\begin{table*}[htbp]
\centering
\caption{Containment Statistics Comparison by Ratio}
\label{tab:containment_compare}
\resizebox{\textwidth}{!}{%
\begin{tabular}{@{} l 
*{7}{S[table-format=3.3] 
     S[table-format=2.3] 
     S[table-format=3.3]} 
@{}}
\toprule
& \multicolumn{21}{c}{Size Ratio} \\
\cmidrule(l){2-22}
\makecell[l]{Method} & 
\multicolumn{3}{c}{1/20} & 
\multicolumn{3}{c}{1/10} & 
\multicolumn{3}{c}{1/5} & 
\multicolumn{3}{c}{1} & 
\multicolumn{3}{c}{5} & 
\multicolumn{3}{c}{10} & 
\multicolumn{3}{c}{20} \\
\cmidrule(lr){2-4} 
\cmidrule(lr){5-7} 
\cmidrule(lr){8-10} 
\cmidrule(lr){11-13} 
\cmidrule(lr){14-16} 
\cmidrule(lr){17-19} 
\cmidrule(lr){20-22}
& {Mean [\%]} & {Std} & {Min [\%]} & 
{Mean [\%]} & {Std} & {Min [\%]} & 
{Mean [\%]} & {Std} & {Min [\%]} & 
{Mean [\%]} & {Std} & {Min [\%]} & 
{Mean [\%]} & {Std} & {Min [\%]} & 
{Mean [\%]} & {Std} & {Min [\%]} & 
{Mean [\%]} & {Std} & {Min [\%]} \\
\midrule
CIRI~\cite{super}         & 98.824 & 8.064 & 0.909 & \textbf{100.000} & \textbf{0.000} & \textbf{100.000} & \textbf{100.000} & \textbf{0.000} & \textbf{100.000} & \textbf{100.000} & \textbf{0.000} & \textbf{100.000} & \textbf{100.000} & \textbf{0.000} & \textbf{100.000} & 99.980 & 0.569 & 83.636 & 96.364 & 10.551 & 20.000 \\
FIRI-Inflation~\cite{firi} & 97.887 & 11.139 & 10.000 & \textbf{100.000} & \textbf{0.000} & \textbf{100.000} & \textbf{100.000} & \textbf{0.000} & \textbf{100.000} & \textbf{100.000} & \textbf{0.000} & \textbf{100.000} & \textbf{100.000} & \textbf{0.000} & \textbf{100.000} & \textbf{100.000} & \textbf{0.000} & \textbf{100.000} & 97.665 & 12.737 & 10.000 \\
FIRI-Shrinkage~\cite{firi}  & 98.343 & 4.477 & 70.909 & 99.960 & 0.525 & 90.909 & 99.996 & 0.083 & 98.182 & \textbf{100.000} & \textbf{0.000} & \textbf{100.000} & 99.998 & 0.042 & 99.091 & 99.955 & 0.606 & 89.091 & 99.349 & 3.045 & 68.182 \\
MACIRI (ours)       & \textbf{100.000} & \textbf{0.000} & \textbf{100.000} & \textbf{100.000} & \textbf{0.000} & \textbf{100.000} & \textbf{100.000} & \textbf{0.000} & \textbf{100.000} & \textbf{100.000} & \textbf{0.000} & \textbf{100.000} & \textbf{100.000} & \textbf{0.000} & \textbf{100.000} & \textbf{100.000} & \textbf{0.000} & \textbf{100.000} & \textbf{100.000} & \textbf{0.000} & \textbf{100.000} \\
\bottomrule
\end{tabular}%
}
\end{table*}

\begin{figure*}[!t]
\centering
\includegraphics[width=\textwidth]{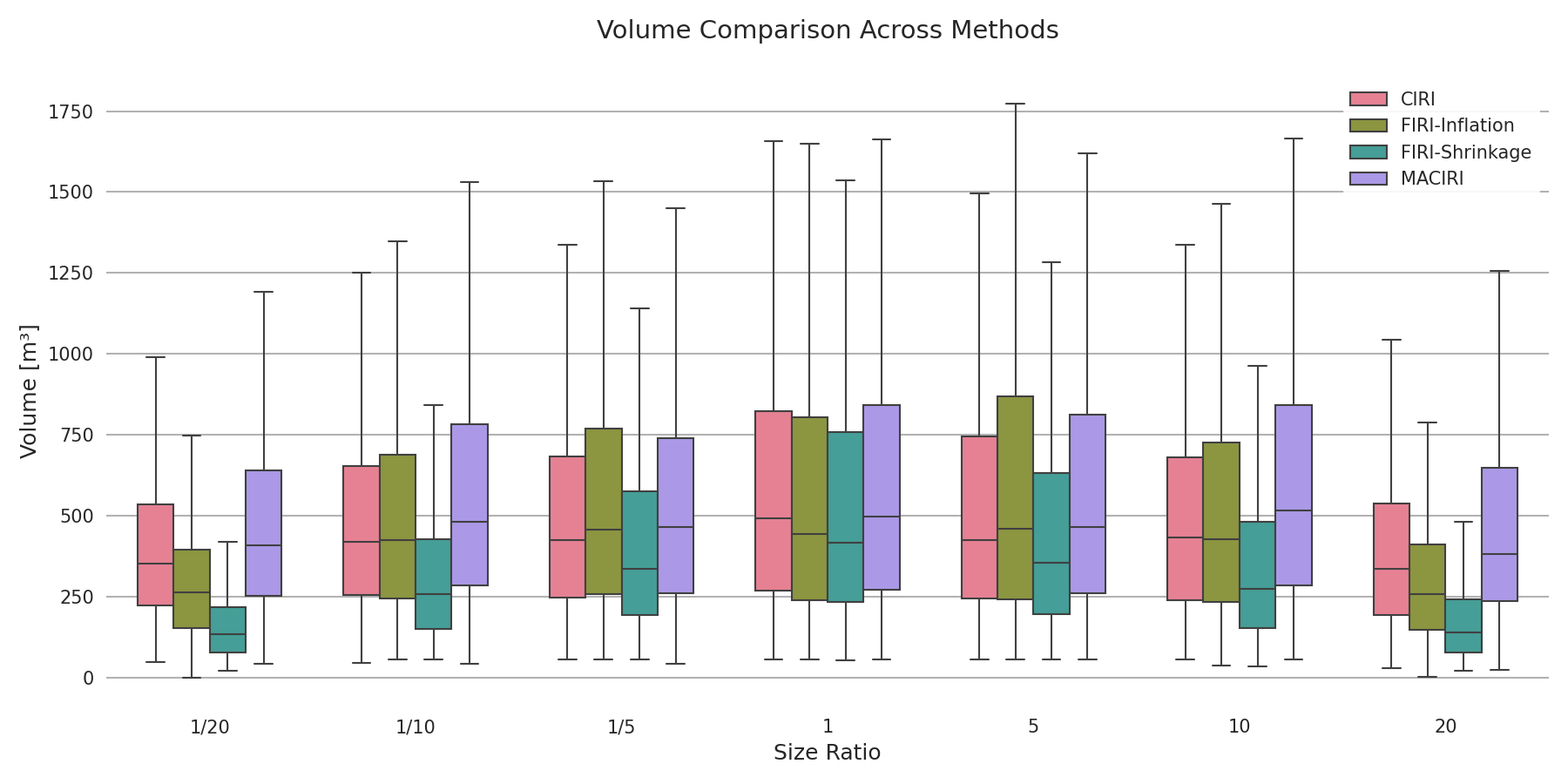}
\caption{Volume comparison of different methods at different size ratios.}
\label{MACIRI_VOL}
\end{figure*}

\begin{figure*}[!t]
\centering
\includegraphics[width=\textwidth]{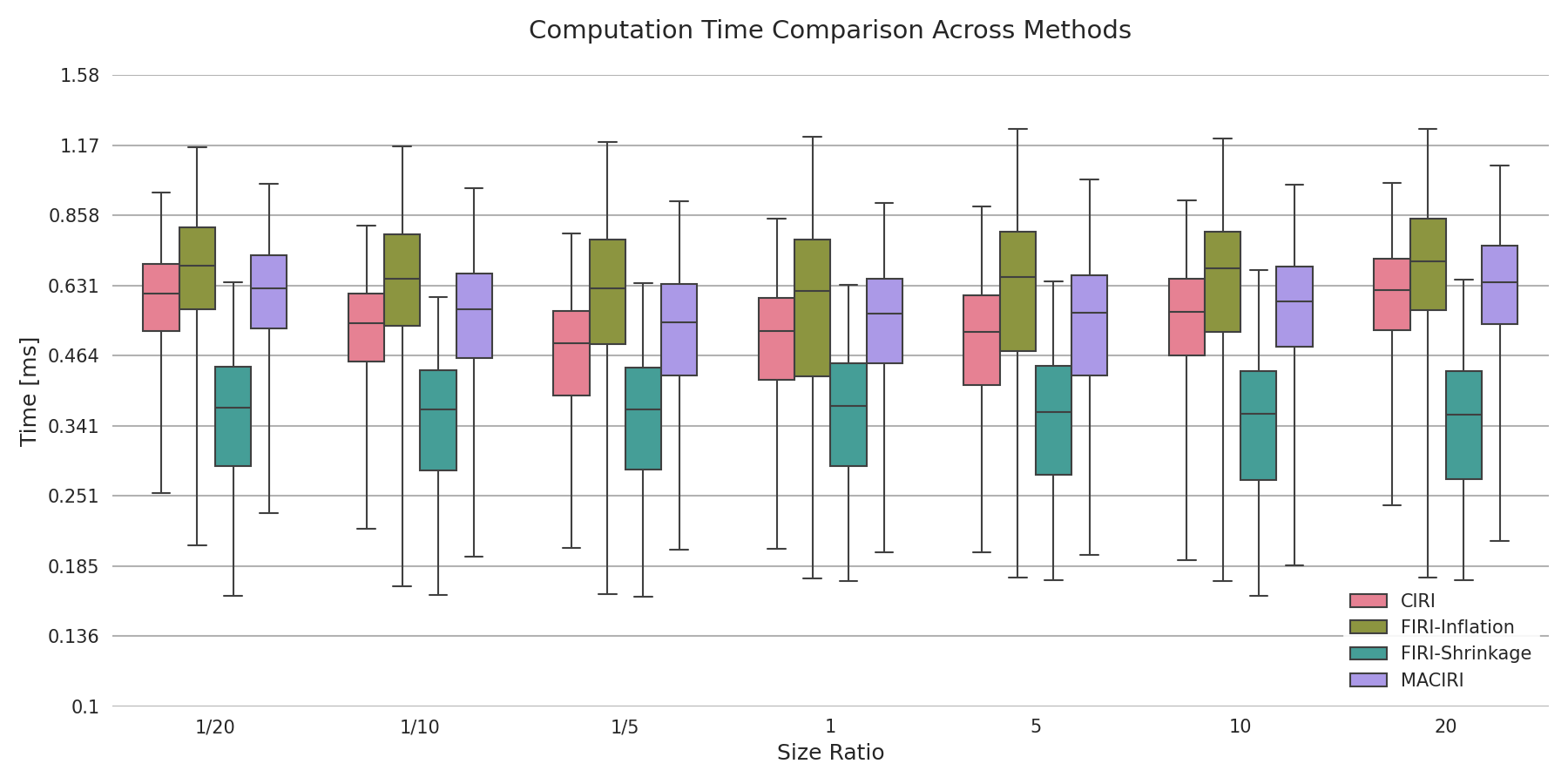}
\caption{Computation time comparison of different methods at different size ratios.}
\label{MACIRI_TIME}
\end{figure*}

We evaluated each method in terms of seed containment rate, computation time, and the volume of the resulting polytope. Regarding seed containment (Table~\ref{tab:containment_compare}), MACIRI consistently achieves 100\% containment across all size ratios. In contrast, the other methods fail to guarantee complete seed inclusion. Specifically, due to the absence of a convex hull–based seed correction strategy, CIRI \cite{super} exhibits poor containment when the ratio is 20, 10, or $\tfrac{1}{20}$. By contrast, FIRI-Inflation \cite{firi} is limited by the coarse resolution of the Nine-Panel inflation; it cannot accurately represent the free space and even overinflates, thereby erroneously including seeds that should not be contained. This overinflation leads to low containment at ratios of 20 and $\tfrac{1}{20}$. For FIRI-Shrinkage \cite{firi}, the shrinkage after iterative convergence naturally excludes seeds, except when the ratio equals 1. Considering the volume of the polytope (Fig.~\ref{MACIRI_VOL}), MACIRI is able to generate substantially larger polytopes when the difference in robot sizes is large. In contrast, the other methods adopt a conservative strategy by using the maximum robot size, which significantly limits their performance under large size differences. With respect to computation time (Fig.~\ref{MACIRI_TIME}), MACIRI requires less time to solve than FIRI-Inflation\cite{firi}. Compared to CIRI \cite{super}, however, the improvement in computation time is relatively modest. In contrast, although MACIRI incurs a longer computation time than FIRI-Shrinkage \cite{firi}, it demonstrates superior performance in other aspects. In the experiment with ratio 20, ablating the multi-size template strategy leads to a substantial decrease in the polytope volume (Table~\ref{tab:volume_ablate}), whereas ablating the correction strategy results in a pronounced reduction in the seed containment rate (Table~\ref{tab:containment_ablate}).

\begin{table*}[h]
\centering
\caption{Volume Analysis Of Ablation Experiments For MACIRI}
\label{tab:volume_ablate}
\resizebox{\textwidth}{!}{
\begin{tabular}{lrrrrrrrr}
\toprule
Method& Mean [m$^3$]  & Min [m$^3$] & 25\% [m$^3$] & 50\% [m$^3$] & 75\% [m$^3$] & Max [m$^3$] \\
\midrule
No Template Policy & 409.907 & 23.097 & 193.75 & 334.931 & 525.028 & 2594.97 \\
Our Full Method & \textbf{499.162}  & \textbf{23.113} & \textbf{235.24} & \textbf{381.719} & \textbf{646.686} & \textbf{3714.06} \\
\bottomrule
\end{tabular}
}
\end{table*}

\begin{table}[h]
\centering
\caption{Containment Rate Analysis Of Ablation Experiments For MACIRI}
\label{tab:containment_ablate}
\begin{tabular}{lrrrrrr}
\toprule
Method & Mean [\%] & Std & Max [\%] & Min [\%]\\
\midrule
No Adjustment Policy & 97.84 & 8.29 & \textbf{100.00} & 21.82\\
Our Full Method & \textbf{100.00} & \textbf{0.00} & \textbf{100.00} & \textbf{100.00}\\
\bottomrule
\end{tabular}
\end{table}

\subsection{Performance of Whole-Body Obstacle Avoidance}
\subsubsection{Performance Across Obstacle Types}
To evaluate the effectiveness of the proposed approach, we design a series of experiments in the MuJoCo simulator \cite{mujoco} under diverse obstacle configurations. Our proposed method, Acetrans, is evaluated against two SOTA baselines, Autotrans~\cite{autotrans} and Impactor~\cite{impactor}. For both Autotrans~\cite{autotrans} and Impactor~\cite{impactor}, we directly employ their publicly available implementations for evaluation. Since Impactor~\cite{impactor} does not provide a replanning mode, and its architecture is closely related to that of Autotrans~\cite{autotrans}, we extend Impactor~\cite{impactor} by incorporating the replanning mechanism of Autotrans~\cite{autotrans} for a fair comparison. The experimental setup is summarized as follows:

\begin{enumerate}
    \item For each obstacle setting, ten maps are randomly generated: one with 70 cylindrical poles and the other with 20 cubic obstacles. Each method is evaluated three times on every map under two environmental conditions: without wind disturbance and with an additional 8\,m/s wind disturbance. Identical map generation parameters are applied across all three methods to ensure fairness.
    \item A quadrotor with a suspended payload system of full length 0.75\,m is required to pass through narrow holes of widths 0.75\,m, 0.70\,m, 0.65\,m, and 0.60\,m. Each method is evaluated ten times for each hole size.
    \item The system is tasked with traversing through arrays of thin pole-like obstacles with radii of 0.05\,m, 0.02\,m, and 0.001\,m. Each method is evaluated ten times for each radius. For Autotrans~\cite{autotrans}, since obstacle avoidance along the cable is enforced by sampling, we consider four configurations with different sampling resolutions: 0.1, 0.05, and 0.001\,m, as well as 0.001\,m combined with an additional 0.4\,m safety margin, which was carefully tuned through repeated experiments by us.

\end{enumerate}

\begin{figure}[b]
\centering
\includegraphics[width=0.40\textwidth]{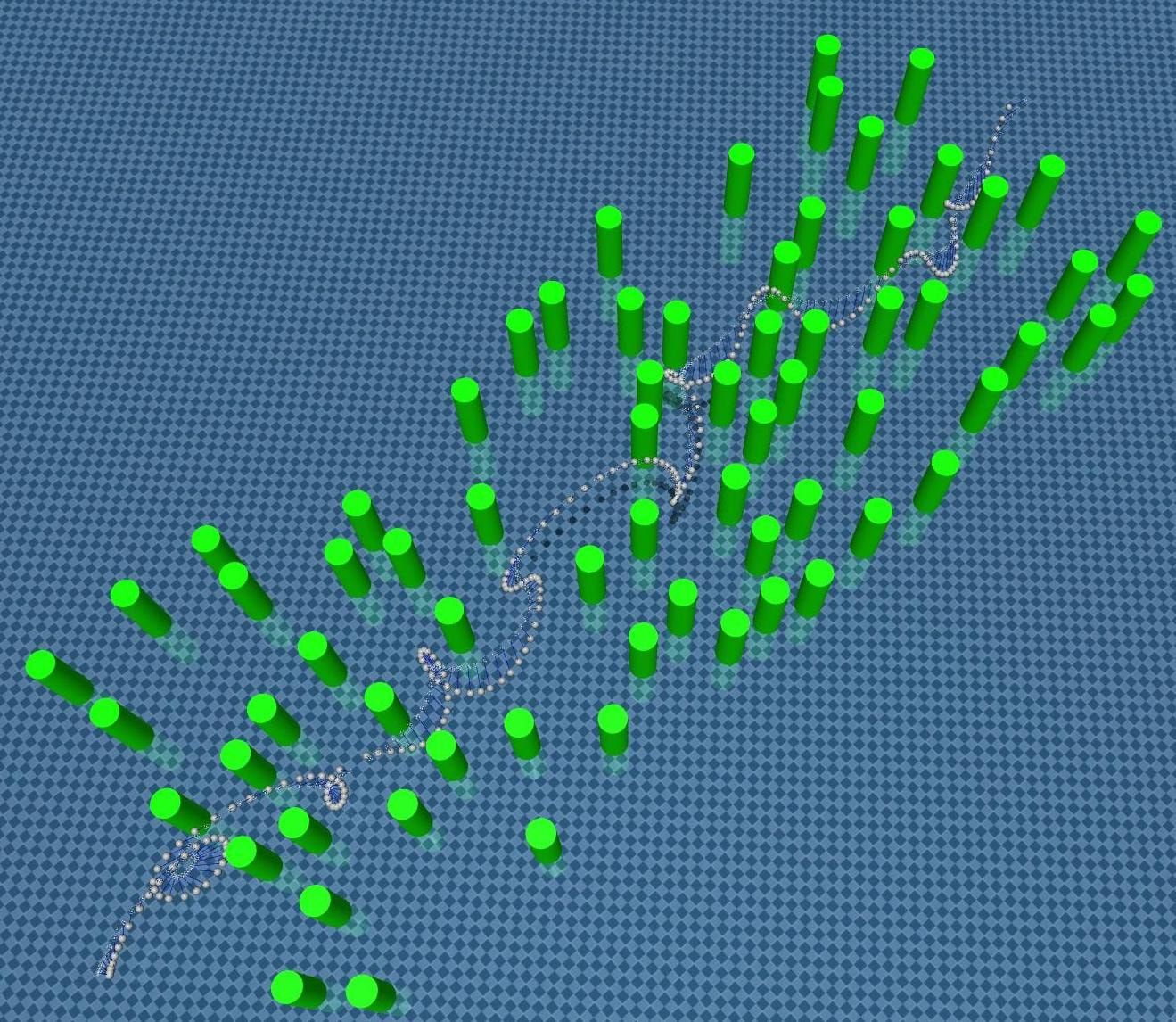}
\caption{Trajectory generated by the proposed planner in pillar-dense scenario.}
\label{fig:Pillar-dense}
\end{figure}

\begin{figure}[tbp]
\centering
\includegraphics[width=0.35\textwidth]{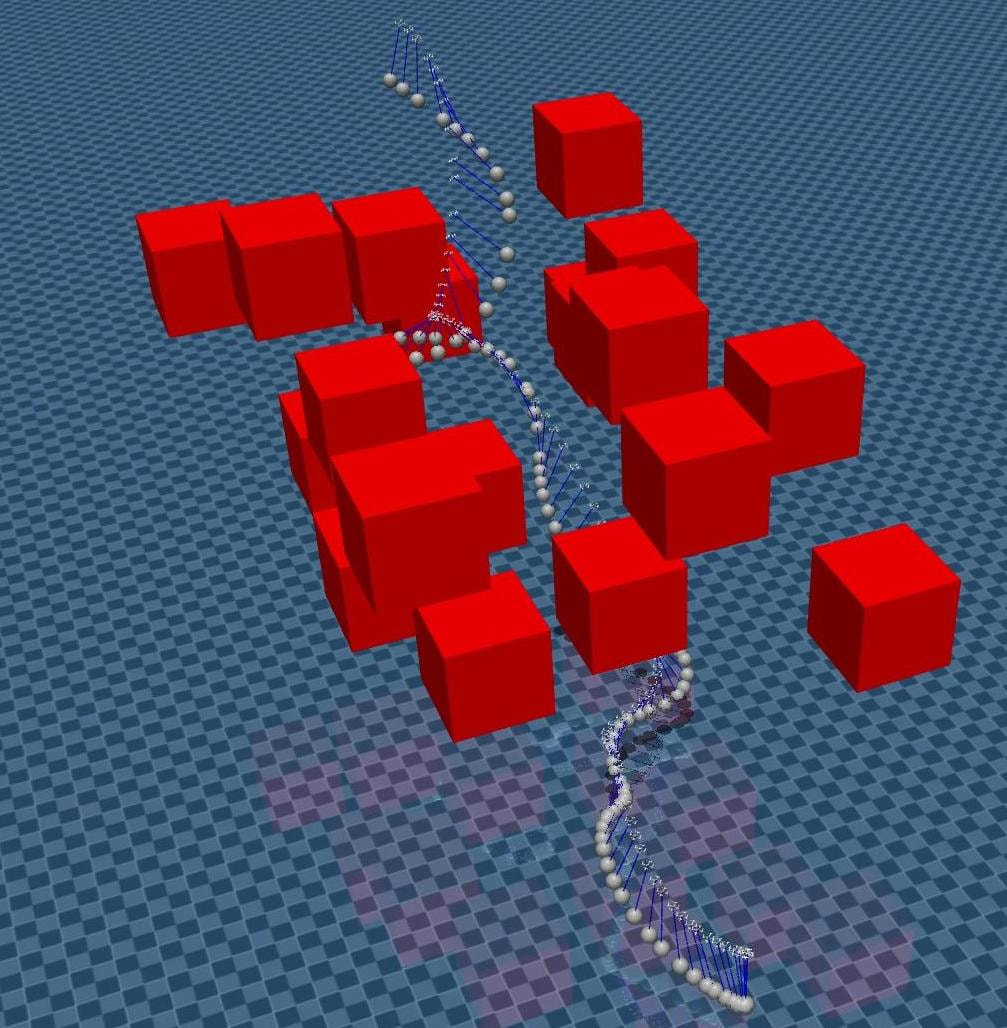}
\caption{Trajectory generated by the proposed planner in cube-dense scenario.}
\label{fig:Cube-dense}
\end{figure}

Tables~\ref{tab:replan compare} report the replanning performance of different methods in randomly generated cluttered maps, including pillar-dense and cube-dense scenarios, both under nominal conditions and with an additional wind disturbance of 8\,m/s. We denote optimization time as Opt, and present both the average and maximum values. The simulation results of the proposed algorithm in the pillar-dense and cube-dense scenarios are shown in Figs.~\ref{fig:Pillar-dense} and \ref{fig:Cube-dense}, respectively.  As illustrated in Tables~\ref{tab:replan compare}, Acetrans achieves optimization efficiency that is one to three orders of magnitude higher than the two baseline methods~\cite{autotrans, impactor}. Moreover, since it does not require the construction of ESDF maps, the CPU usage of Acetrans is significantly lower compared with both baselines. Benefiting from safer obstacle-avoidance guarantees and the substantial improvement in planning efficiency provided by the corridor-based formulation (Section~\ref{sec:property_corr}), Acetrans attains higher task success rates(Succ.) and superior obstacle-avoidance performance. Furthermore, by explicitly accounting for external forces in the differential flatness framework (Section~\ref{sec:flatness}), Acetrans is able to maintain robust performance even under strong wind disturbances.

\begin{table*}[h]
\centering
\caption{Replanning Performance Comparison of Three Methods}
\label{tab:replan compare}
\resizebox{\textwidth}{!}{
\begin{tabular}{lcccccc}
\toprule
Scenario & Method & Opt(avg/max) [ms] & ESDF Time [ms]& Succ. [\%] & Usage(CPU/Memory) [\%] \\
\midrule
\multirow{3}{*}{Pillars} 
& Autotrans~\cite{autotrans}& 24.151/670.711 &65.497 & 16 & 83.8/5.7\\
& Impactor~\cite{impactor}& 1560.094/5016.000 &60.915 & 0 &83.1/14.5\\
& Acetrans (ours)& \textbf{1.589/3.887} &- & \textbf{100} & \textbf{8.0/3.9}\\
\midrule
\multirow{3}{*}{Pillars + 8m/s wind} 
& Autotrans~\cite{autotrans}& 7.394/78.007 &64.903 & 0 & 84.2/6.1\\
& Impactor~\cite{impactor}& 1300.753/2012.990 &66.133 & 0 & 85.4/18.9\\
& Acetrans (ours)& \textbf{1.765/8.807} &- & \textbf{83} & \textbf{8.3/3.8}\\

\midrule
\multirow{3}{*}{Cubes} 
& Autotrans~\cite{autotrans}& 13.574/291.939  &54.809 & 56 & 75.4/24.0\\
& Impactor~\cite{impactor}& 3082.176/18759.100 &56.215 & 23 &76.0/29.1\\
& Acetrans (ours)&\textbf{0.464/1.628} &- & \textbf{100} & \textbf{11.1/14.4}\\

\midrule
\multirow{3}{*}{Cubes + 8m/s wind} 
& Autotrans~\cite{autotrans}&9.227/261.501 &56.490 & 10 & 74.8/23.7\\
& Impactor~\cite{impactor}& 1257.127/1542.770 &55.852 & 0 & 72.1/28.8\\
& Acetrans (ours)&\textbf{0.454/1.481} &- & \textbf{100} & \textbf{12.2/14.6}\\

\bottomrule
\end{tabular}
}
\end{table*}

% \begin{table*}[h]
% \centering
% \caption{Replanning Performance Comparison of Three Methods}
% \label{tab:replan compare}
% \resizebox{\textwidth}{!}{
% \begin{tabular}{lcccccc}
% \toprule
% Method & Scenario & Opt(avg/max) [ms] & ESDF Time [ms]& Succ. [\%] & Usage(CPU/Memory) [\%] \\
% \midrule
% \multirow{4}{*}{Acetrans} 
% & Pillars& 1.589/3.887 &- & 100 & 8.0/3.9\\
% & Pillars + 8m/s wind& 1.765/8.807 &- & 83 & 8.3/3.8\\
% & Cubes& 0.464/1.628 &- & 100 & 11.1/14.4\\
% & Cubes + 8m/s wind& 0.454/1.481 &- & 100 & 12.2/14.6\\
% \midrule
% \multirow{4}{*}{Autotrans~\cite{autotrans}}
% & Pillars& 24.151/670.711 &65.497 & 16 & 83.8/5.7\\
% & Pillars + 8m/s wind& 7.394/78.007 &64.903 & 0 & 84.2/6.1\\
% & Cubes& 13.574/291.939  &54.809 & 56 & 75.4/24.0\\
% & Cubes + 8m/s wind& 9.227/261.501 &56.490 & 10 & 74.8/23.7\\
% \midrule
% \multirow{4}{*}{Impactor~\cite{impactor}}
% & Pillars& 1560.094/5016.000 &60.915 & 0 &83.1/14.5\\
% & Pillars + 8m/s wind& 1300.753/2012.990 &66.133 & 0 & 85.4/18.9\\
% & Cubes& 3082.176/18759.100 &56.215 & 23 &76.0/29.1\\
% & Cubes + 8m/s wind& 1257.127/1542.770 &55.852 & 0 & 72.1/28.8\\
% \bottomrule
% \end{tabular}
% }
% \end{table*}

Table~\ref{tab:time distrib} reports the planning time distribution of Acetrans. We denote the SFC generation time as SFC, the initialization time as Init, and the total time as Total, and report both the average and maximum values.
In pillar-dense scenarios, wind disturbance leads to a moderate increase in computation time.
Across all cases, the majority of time is spent on SFC generation and optimization, 
and the overall runtime remains well within real-time requirements, 
demonstrating both efficiency and robustness of Acetrans.

\begin{table*}[h]
\centering
\caption{Planning Time Distribution Analysis for Acetrans}
\label{tab:time distrib}
\resizebox{\textwidth}{!}{
\begin{tabular}{lcccc}
\toprule
Scenario & SFC(avg/max) [ms] & Init(avg/max) [ms] & Opt(avg/max) [ms] & Total(avg/max) [ms] \\
\midrule
Pillars & 1.591/4.603 & 0.061/0.190 & 1.589/3.887 & 3.241/7.442 \\
Pillars + 8m/s wind & 1.723/4.983 & 0.067/0.307 & 1.765/8.807 & 3.556/10.110 \\
Cubes & 1.158/4.239 & 0.037/0.175 & 0.464/1.628 & 1.660/5.001 \\
Cubes + 8m/s wind & 1.061/3.955 & 0.033/0.170 & 0.454/1.481 &  1.549/5.412 \\
\bottomrule
\end{tabular}
}
\end{table*}

\begin{figure}[!t]
\centering
\includegraphics[width=0.48\textwidth]{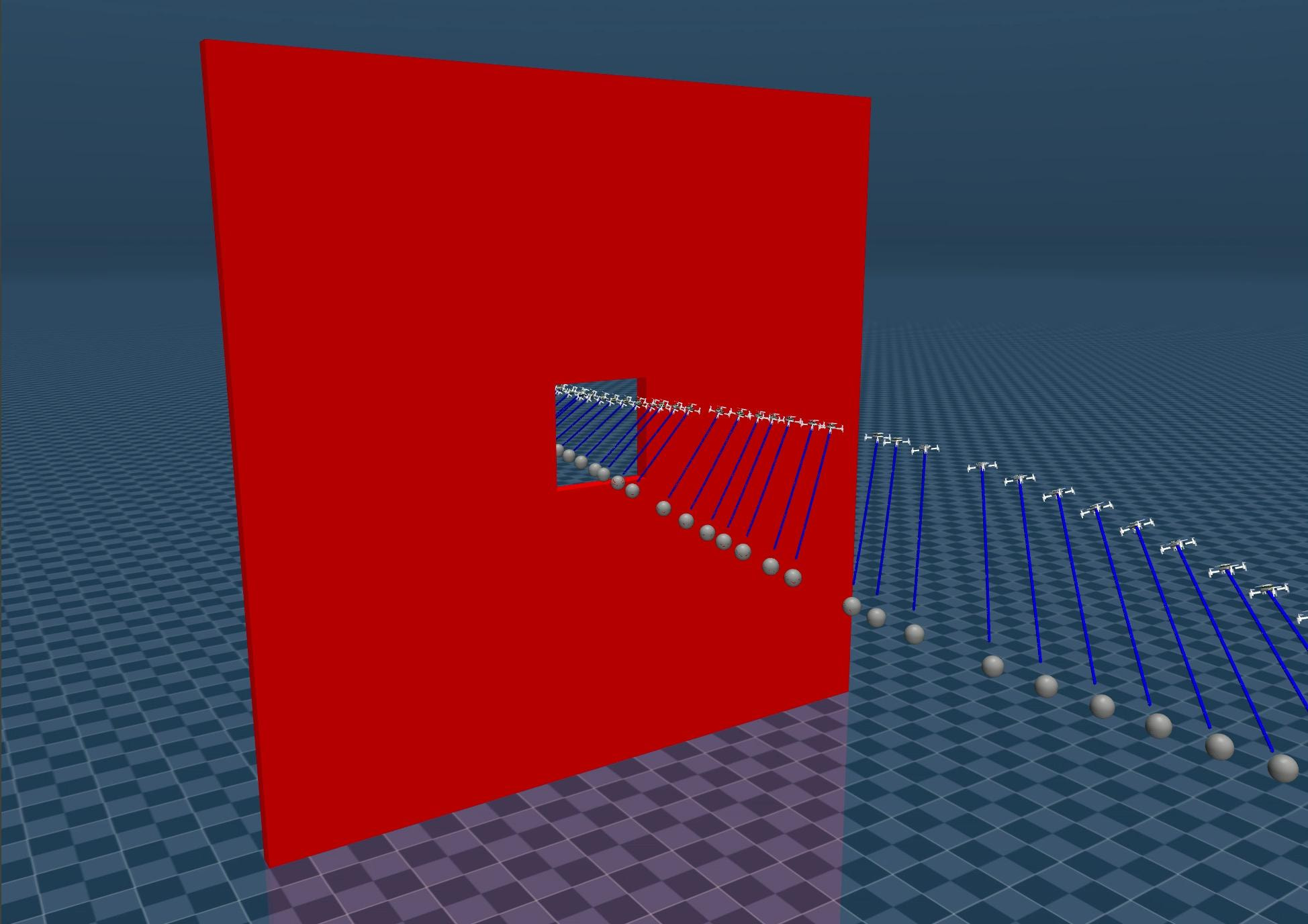}
\caption{Trajectory generated by the proposed planner for a quadrotor with a suspended payload traversing a narrow gap whose width does not exceed the vehicle length.}
\label{fig:narrow_gap}
\end{figure}

Table~\ref{tab:cq_comparison} summarizes the performance of different methods when the quadrotor with a suspended payload is required to traverse narrow gaps whose widths are less than or equal to its full length, while the simulation results of the proposed algorithm are illustrated in Fig.~\ref{fig:narrow_gap}. The results show that Acetrans consistently achieves both efficient optimization and a high success rate in planning and control. In contrast, Autotrans~\cite{autotrans} struggles to generate feasible solutions as the gap narrows, leading to a sharp drop in overall success. Although Impactor demonstrates improved flexibility at the planning stage by considering cable bending, this approach introduces a discrepancy between planned and executable trajectories. As a consequence, even when feasible plans are obtained in narrower gaps (e.g., at 0.65\,m and 0.6\,m), the controller frequently fails to track them reliably, resulting in collisions. This highlights the importance of jointly ensuring both planning feasibility and control robustness in suspended payload transportation.

\begin{table}[h]
	\centering
	\caption{Performance Comparison Across Methods With Different Holes}
	\label{tab:cq_comparison}
        \resizebox{\linewidth}{!}{
	\begin{tabular}{lcccc}
		\toprule
		Size [m] & Method & Opt(avg/max) [ms]& Succ.(Plan/Control) [\%] \\
		\midrule   
            \multirow{3}{*}{0.75} 
		& Autotrans~\cite{autotrans} & 24.702/68.421 & 30/30\\
		& Impactor~\cite{impactor} & 2753.612/2873.450 & 100/100\\
            & Acetrans (ours) & \textbf{0.952/1.373} & \textbf{100/100}\\
            
		\midrule
            \multirow{3}{*}{0.70} 
		& Autotrans~\cite{autotrans} & 21.255/66.349 & 0/0\\
		& Impactor~\cite{impactor} & 2605.776/2777.530 & 100/100\\
            & Acetrans (ours) & \textbf{1.099/1.766} & \textbf{100/100}\\

		\midrule
            \multirow{3}{*}{0.65} 
		& Autotrans~\cite{autotrans} & 30.023/ 64.880 & 0/0\\
		& Impactor~\cite{impactor} & 1552.564/1704.830 & 100/0\\
            & Acetrans (ours) & \textbf{0.897/1.396} & \textbf{100/100}\\

		\midrule
            \multirow{3}{*}{0.60} 
		& Autotrans~\cite{autotrans} & 69.001/282.458 & 0/0\\
		& Impactor~\cite{impactor} & 1553.310/1657.600 & 100/0 \\
            & Acetrans (ours) & \textbf{1.193/1.579} & \textbf{100/100}\\

		\bottomrule
	\end{tabular}}
\end{table}

\begin{figure}[tbp]
\centering
\includegraphics[width=0.48\textwidth]{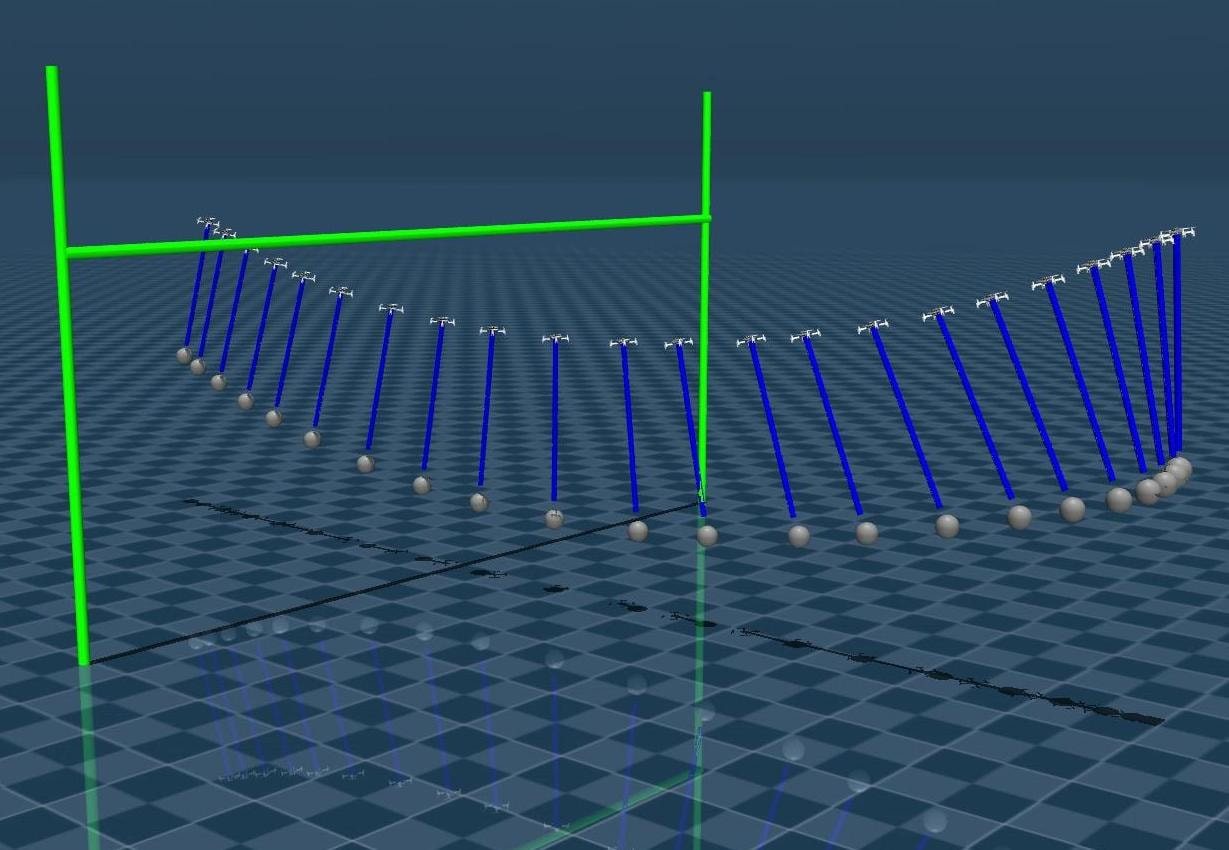}
\caption{Trajectory generated by the proposed planner in an environment with slender pole-like obstacles.}
\label{fig:pole_obstacles}
\end{figure}

Table~\ref{tab:tp_comparison} reports the obstacle-avoidance performance against slender pole-like obstacles with varying radii., while the simulation results of the proposed algorithm are illustrated in Fig.~\ref{fig:pole_obstacles}.
\,For Impactor~\cite{impactor}, since whole-body obstacle avoidance is not considered, the success rate remains zero in all tests, and the computational time is significantly higher than that of the other two methods. 
For Autotrans~\cite{autotrans}, the solving time increases as the cable sampling resolution becomes finer. At low resolutions, the success rate is zero, and even at a sampling resolution of 0.001\,m, the method still fails to guarantee whole-body safety, resulting in collisions with slender obstacles. After extensive parameter tuning, we found that adding an additional safety margin of 0.4\,m on top of the 0.001\,m sampling resolution can improve success rates; however, this comes at the cost of a substantial increase in computation time. 
In contrast, our proposed method, Acetrans, consistently outperforms both baselines in terms of success rate and computational efficiency. Remarkably, it achieves $100\%$ success even when traversing extremely thin obstacles with a diameter of 0.001\,m, while maintaining low computational overhead. 
This advantage arises because Acetrans does not rely on discretizing the suspended cable for obstacle avoidance. 
Unlike Autotrans~\cite{autotrans}, which requires increasing the cable sampling resolution to handle thin obstacles—thereby significantly inflating optimization time, Acetrans maintains nearly constant efficiency and ensures reliable whole-body safety regardless of obstacle size.

After comparing with the two baseline methods, we further conduct an ablation study on the proposed trajectory initialization strategy. As shown in Table~\ref{tab:ablation_init}, the ablated version adopts an initialization method of \cite{gcopter}, which only pursues the shortest path. As indicated in the table, our initialization method significantly improves the overall task success rate, since the L-BFGS-based trajectory optimizer strongly depends on a well-conditioned initial solution. By ensuring that both the initial quadrotor and payload positions are located within the corridor, our method enables feasible planning in a wider range of scenarios.

\begin{table}[h]
\centering
\caption{Ablation on trajectory initialization. Planning success rates across scenarios comparing the Only Shortest baseline~\cite{gcopter} with our initialization method.}
\label{tab:ablation_init}
\begin{tabular}{llcc}
\toprule
Category & Size [m] & Only Shortest~\cite{gcopter} [\%] & Ours [\%] \\
\midrule
Pillars & --   & 6.6   & \textbf{100.0} \\
\hline
Cubes   & --   & 10.0  & \textbf{100.0} \\
\hline
\multirow{3}{*}{Pole}    & 0.05 & 0.0   & \textbf{100.0} \\
        & 0.02 & 0.0   & \textbf{100.0} \\
        & 0.001& 0.0   & \textbf{100.0} \\
\hline
\multirow{3}{*}{Hole}    & 0.75 & \textbf{100.0} & \textbf{100.0} \\
        & 0.70 & \textbf{100.0} & \textbf{100.0} \\
        & 0.65 & 0.0   & \textbf{100.0} \\
        & 0.60 & 0.0   & \textbf{100.0} \\
\bottomrule
\end{tabular}
\end{table}

% \begin{table}[h]
% \caption{Planning Performance Across Initialization Methods}
% \label{tab:ablation_init}
% \resizebox{0.48\textwidth}{!}{
% \begin{tabular}{lllc}
% \hline
% Category & Size[m] & Method & Succ. [\%] \\
% \hline
% \multirow{2}{*}{Pillars} & -- & Only Shortest~\cite{gcopter} & 6.6 \\
%         & -- & Our Full Method & \textbf{100.0} \\
% \hline
% \multirow{2}{*}{Cubes}   & -- & Only Shortest~\cite{gcopter} & 10.0 \\
%         & -- & Ours & \textbf{100.0} \\
% \hline
% \multirow{6}{*}{Pole} 
%     & \multirow{2}{*}{0.05} & Only Shortest~\cite{gcopter} & 0.0 \\
%     &                         & Ours & \textbf{100.0} \\
%     \cline{2-4}
%     & \multirow{2}{*}{0.02} & Only Shortest~\cite{gcopter} & 0.0 \\
%     &                         & Ours & \textbf{100.0} \\
%     \cline{2-4}
%     & \multirow{2}{*}{0.001} & Only Shortest~\cite{gcopter} & 0.0 \\
%     &                          & Ours & \textbf{100.0} \\
% \hline
% \multirow{8}{*}{Hole}
%     & \multirow{2}{*}{0.75} & Only Shortest~\cite{gcopter} & \textbf{100.0} \\
%     &                          & Ours & \textbf{100.0} \\
%     \cline{2-4}
%     & \multirow{2}{*}{0.70} & Only Shortest~\cite{gcopter} & \textbf{100.0} \\
%     &                          & Ours & \textbf{100.0} \\
%     \cline{2-4}
%     & \multirow{2}{*}{0.65} & Only Shortest~\cite{gcopter} & 0.0 \\
%     &                          & Ours & \textbf{100.0} \\
%     \cline{2-4}
%     & \multirow{2}{*}{0.60} & Only Shortest~\cite{gcopter} & 0.0 \\
%     &                          & Ours & \textbf{100.0} \\
% \hline
% \end{tabular}
% }
% \end{table}

\begin{table}[h]
\centering
\caption{Performance Comparison Across Methods With Different Thin Poles.}
\label{tab:tp_comparison}
\resizebox{0.48\textwidth}{!}{
\begin{tabular}{l l c c c}
\toprule
Radius [m] & Method & Sampling length [m] & Opt (avg/max) [ms] & Succ. [\%] \\
\midrule
\multirow{6}{*}{0.05}
  & \multirow{4}{*}{Autotrans~\cite{autotrans}}
      & 0.1   & 2.766/7.381   & 0 \\
  &   & 0.05  & 3.821/11.287  & 0 \\
  &   & 0.001 & 13.713/26.015 & 0 \\
  &   & \makecell[c]{0.001\\\footnotesize(+0.4 m safety margin)} & 44.005/94.301 & \textbf{100} \\
  \cline{2-5}
  & Impactor~\cite{impactor} & -- & 1815.352/1913.890 & 0 \\
    \cline{2-5}
    & Acetrans (ours) & -- & \textbf{0.841/1.543} & \textbf{100} \\

\midrule
\multirow{6}{*}{0.02}
  & \multirow{4}{*}{Autotrans~\cite{autotrans}}
      & 0.1   & 3.166/4.616   & 0 \\
  &   & 0.05  & 3.657/8.897   & 0 \\
  &   & 0.001 & 11.762/16.292 & 0 \\
  &   & \makecell[c]{0.001\\\footnotesize(+0.4 m safety margin)} & 55.268/119.047 & \textbf{100} \\
  \cline{2-5}
  & Impactor~\cite{impactor} & -- & 1881.651/1941.220 & 0 \\
    \cline{2-5}
    & Acetrans (ours) & -- & \textbf{0.837/1.428} & \textbf{100} \\

\midrule
\multirow{6}{*}{0.001}
  & \multirow{4}{*}{Autotrans~\cite{autotrans}}
      & 0.1   & 3.090/7.336   & 0 \\
  &   & 0.05  & 3.842/10.191  & 0 \\
  &   & 0.001 & 14.021/22.383 & 0 \\
  &   & \makecell[c]{0.001\\\footnotesize(+0.4 m safety margin)} & 37.524/61.491 & 90 \\
  \cline{2-5}
  & Impactor~\cite{impactor} & -- & 1787.168/1824.610 & 0 \\
    \cline{2-5}
  & Acetrans (ours) & -- & \textbf{0.767/1.120} & \textbf{100} \\
\bottomrule
\end{tabular}}
\end{table}

\subsubsection{Performance of Cable-Bending Obstacle Avoidance}

\begin{figure}[!t]
\centering
\includegraphics[width=0.48\textwidth]{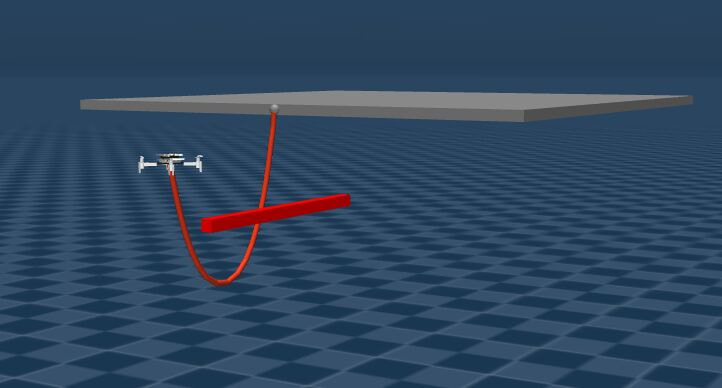}
\caption{Simulation setup for evaluating cable-bending obstacle avoidance.}
\label{fig:cable_bending_env}
\end{figure}

To evaluate the performance of cable-bending obstacle avoidance, we design a scenario in which the payload is fixed at a specific point to simulate the loading and unloading process, with the quadrotor hovering nearby awaiting the completion of the task, and the simulation environment is illustrated in Fig.~\ref{fig:cable_bending_env}. During the process, random wind disturbances and dynamic obstacles approaching the bend cable are introduced. For the wind disturbances, three sets of conditions are applied, each generating wind along different axes with amplitudes of 1, 2, and 3, respectively. These disturbances are time-varying, which in turn causes the shape of the bend cable to change dynamically, significantly increasing the complexity of the experiment. For dynamic obstacles, we set their velocities to 0.5, 1.0, and 1.5\,m/s. The results of the experiment, as shown in Table~\ref{tab:cable-bending obstacle avoidance}, demonstrate that our algorithm is capable of functioning effectively under random, time-varying wind disturbances and dynamic obstacles.

\begin{table}[!b]
\centering
\caption{Success Rate of Cable Bending to Avoid Obstables}
\label{tab:cable-bending obstacle avoidance}
\begin{tabular}{ccc}
\toprule
Wind Speed [m/s] & Obstable Speed [m/s] & Succ. [\%] \\
\midrule
\multirow{3}{*}{1} & 0.5 & 100 \\
                   & 1.0 & 100 \\
                   & 1.5 & 100 \\
\hline
\multirow{3}{*}{2} & 0.5 & 100 \\
                   & 1.0 & 100 \\
                   & 1.5 & 100 \\
\hline
\multirow{3}{*}{3} & 0.5 & 100 \\
                   & 1.0 & 100 \\
                   & 1.5 & 100 \\
\bottomrule
\end{tabular}
\end{table}

\subsection{Indoor experiment}
We conducted our experiments in an indoor environment measuring 14 m in length, 5 m in width, and 3.5 m in height. Within this space, we arranged three-dimensional obstacles, including upright cylindrical columns and low ground-level boxes. We set the initial point at position (0,0) and the goal point at (12.5$m$,0). To reach the goal, the trajectory must pass through this region filled with complex obstacles.Two experimental conditions were designed—one without wind disturbance and the other with wind disturbance (with an average wind speed of 3.5 m/s)—to validate the overall performance of our perception, planning, and control algorithmic system.

\begin{figure}[!t]
\centering
\includegraphics[width=0.5
\textwidth]{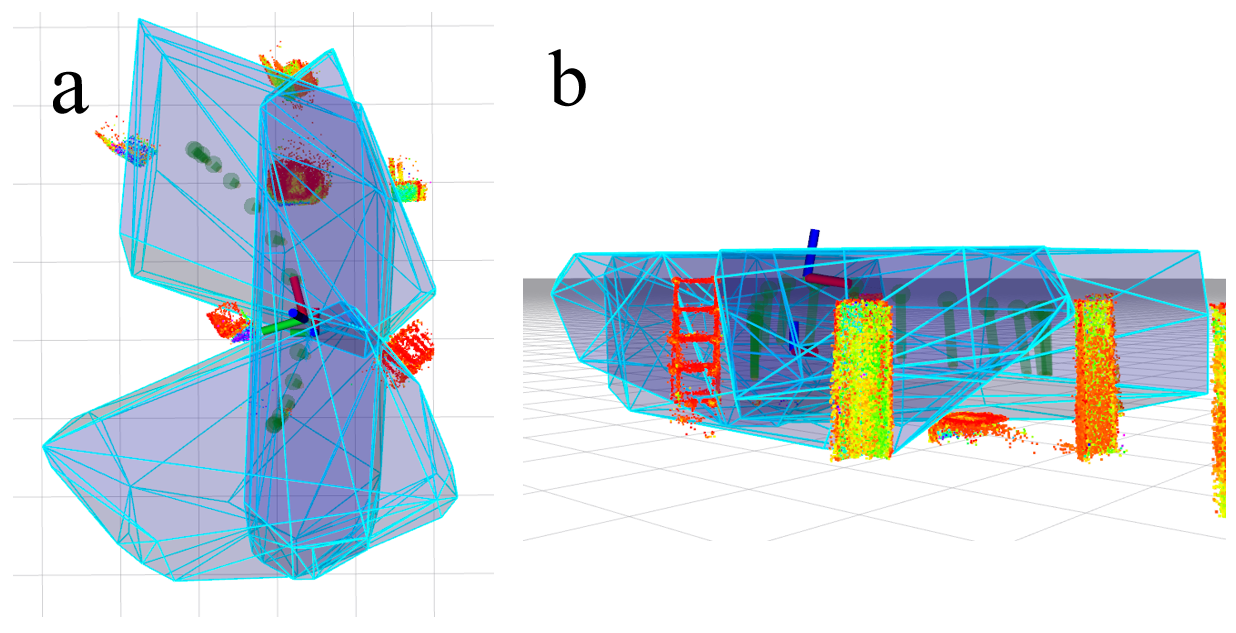}
\caption{Visualization of the indoor experimental corridor and flight trajectories.(a) Top view and (b) front view. The coordinate frames represent the UAV and the autonomously perceived payload. The green slender cylinders denote the trajectories of the UAV–payload system, while the blue convex hulls indicate the corridors generated by MACIRI.}
\label{Visual rviz}
\end{figure}

\begin{figure}[!]
\centering
\includegraphics[width=0.4
\textwidth]{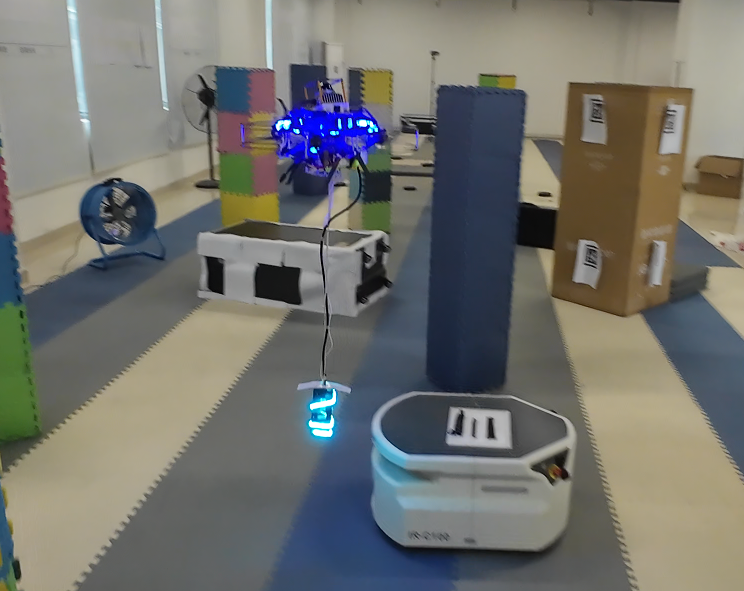}
\caption{Experimental snapshot demonstrating the trajectory planning and control performance without wind disturbance.}
\label{indoor no wind}
\end{figure}

\begin{figure}[!t]
\centering
\includegraphics[width=0.4
\textwidth]{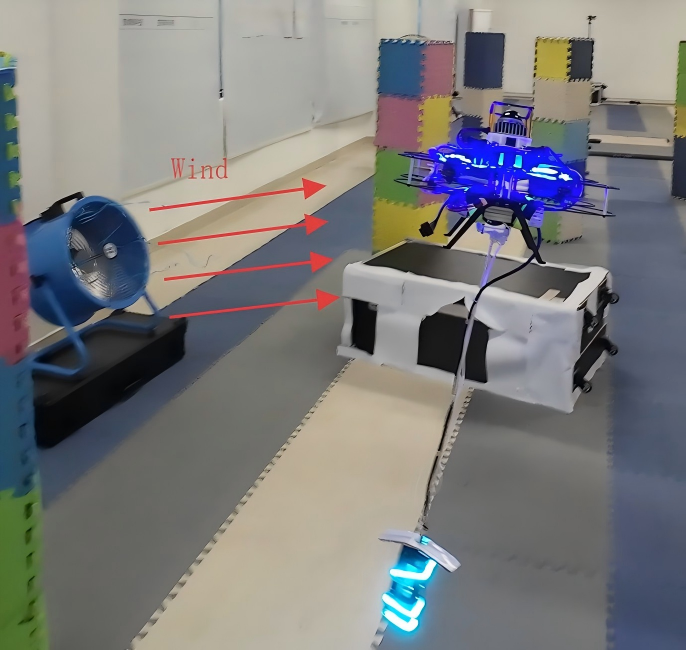}
\caption{Experimental snapshot demonstrating the trajectory planning and control performance under wind disturbance.}
\label{indoor wind}
\end{figure}

The experimental snapshot under no-wind conditions is shown in Fig.  \ref{indoor no wind}, while the snapshot under wind conditions is presented in Fig. \ref{indoor wind}. The rviz visualization is illustrated in Fig. \ref{Visual rviz}.

\subsection{Outdoor experiment}

\subsubsection{Outdoor forest experiment}
We selected a forested environment to conduct the outdoor experiments. To verify that our system is capable of performing transportation tasks at night, we additionally carried out experiments under nighttime conditions.

\begin{figure}[!t]
\centering
\includegraphics[width=0.3
\textwidth]{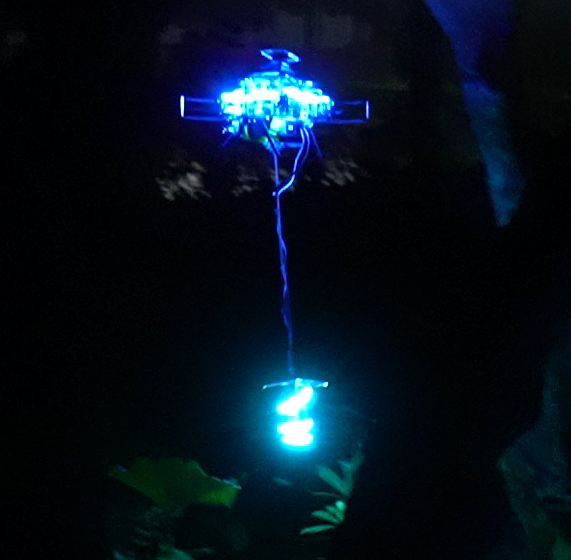}
\caption{Experimental snapshot demonstrating the trajectory planning and control performance during the nighttime outdoor experiment.}
\label{picture:nighttime forest}
\end{figure}

\begin{figure}[!t]
\centering
\includegraphics[width=0.4
\textwidth]{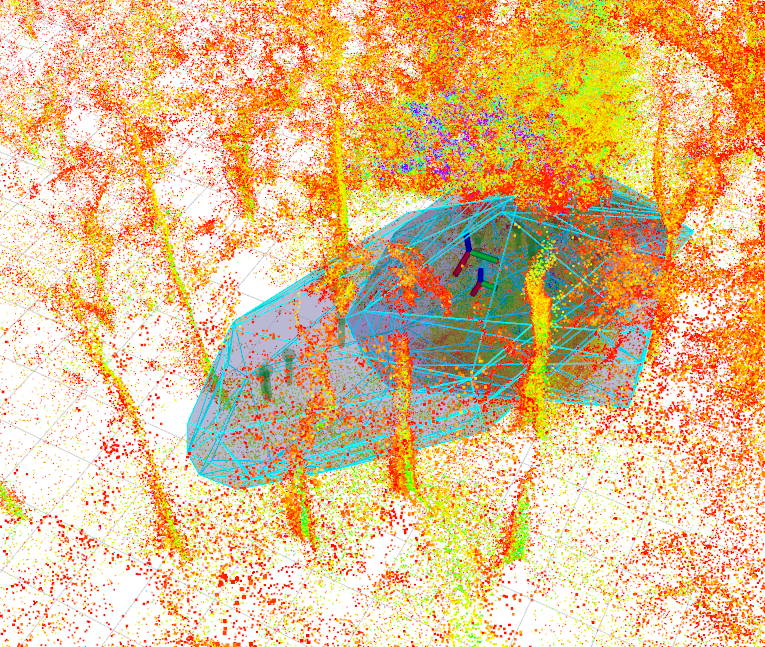}
\caption{The rviz visualization of the outdoor experiment.}
\label{picture:nighttime forest rviz}
\end{figure}

The snapshot of the nighttime outdoor experiment is shown in Fig. \ref{picture:nighttime forest}, and the rviz visualization of the outdoor experiment is presented in Fig. \ref{picture:nighttime forest rviz}.Under both wind and no-wind conditions, the UAV–payload system was able to agilely avoid all obstacles and successfully reach the target destination.
Furthermore, the UAV–payload system was capable of traversing the forest without collisions, completing a looped trajectory of up to 25 m before returning to the starting point.

\subsubsection{Obstacle avoidance experiment with a slender rope}
\begin{figure}[!t]
\centering
\includegraphics[width=0.4
\textwidth]{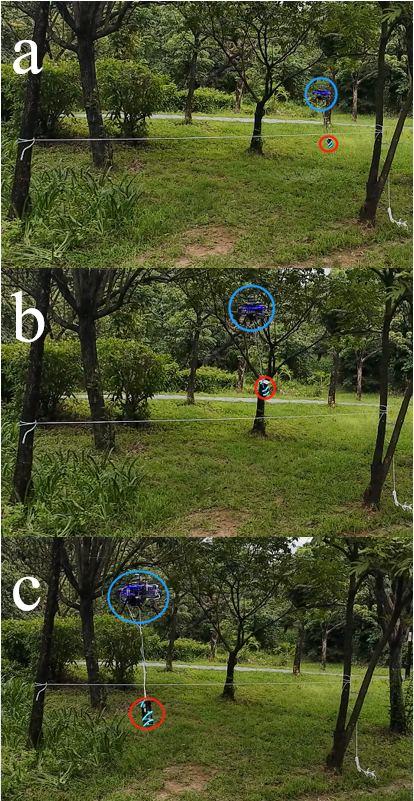}
\caption{Obstacle-avoidance experiment with a slender rope: (a) before crossing the rope, (b) crossing above the rope, and (c) after crossing the rope.}
\label{picture:slender rope}
\end{figure}

We designed an experiment in which a slender rope with a thickness of 4 mm was placed across the flight path of the UAV as an obstacle. The experimental procedure is illustrated in Fig. \ref{picture:slender rope}.The UAV was able to perceive the 4-mm-thick rope, promptly replan its trajectory, and agilely fly over the slender obstacle.

\section{Conclusion}
This paper introduced \textbf{Acetrans}, an \textbf{A}utonomous \textbf{C}orridor-based and \textbf{E}fficient UAV suspended \textbf{trans}port system. In contrast to prior approaches, Acetrans explicitly addresses the challenges of whole-body perception, planning, and control for UAV–payload systems operating in cluttered and dynamic environments.

We proposed a LiDAR–IMU based perception module that jointly estimates both payload pose and cable shape under taut and bent conditions. This capability enables accurate whole-body state estimation and real-time filtering of cable point clouds, ensuring robust mapping and localization. Building on this perception layer, we developed the \textbf{M}ulti-size-\textbf{A}ware \textbf{C}onfiguration-space \textbf{I}terative \textbf{R}egional \textbf{I}nflation (MACIRI) algorithm, which generates safe flight corridors tailored to varying UAV and payload geometries, thereby expanding the usable free space while guaranteeing complete seed containment.

On the planning side, a spatio-temporal, corridor-constrained trajectory optimization method was introduced to produce dynamically feasible and collision-free trajectories. By leveraging the geometric properties of convex corridors, our approach ensures whole-body safety, meaning that both the UAV, payload, and the connecting cable remain entirely within safe regions throughout the entire trajectory. Finally, we designed a NMPC augmented with cable-bending constraints to achieve robust obstacle avoidance under disturbances and significant cable deformation.

Comprehensive simulations and real-world experiments demonstrated that Acetrans achieves superior perception accuracy, planning efficiency, and control robustness compared to state-of-the-art methods. By explicitly addressing whole-body safety, Acetrans enables reliable UAV suspended-load transportation in complex environments. Future work will explore cooperative multi-UAV systems, adaptive planning under dynamic obstacles, and the integration of tri-axial force sensors to further enhance perception accuracy and robustness.


\bibliography{references}

\newpage

\end{document}